\documentclass[conference]{IEEEtran}
\IEEEoverridecommandlockouts
\usepackage{cite}
\usepackage{amsmath,amssymb,amsfonts}
\usepackage{algorithmic}
\usepackage{graphicx}
\usepackage{textcomp}
\usepackage{xcolor}
\usepackage{comment}
\usepackage{hyperref}
\usepackage{booktabs}
\usepackage{multirow}
\usepackage{tabularx}

\def\BibTeX{{\rm B\kern-.05em{\sc i\kern-.025em b}\kern-.08em
    T\kern-.1667em\lower.7ex\hbox{E}\kern-.125emX}}
\begin{document}

\title{Cross-Lingual Transferability of Training Data Extraction Attacks to Recover Memorized PII
}


\author{\IEEEauthorblockN{Alexandru Nazare$^\dagger$, Agnese Profico$^\dagger$, Nicolò Vania$^\dagger$, Elena Di Croce$^\dagger$, Daria Caramanica$^\dagger$,}
\IEEEauthorblockN{Davide Venditti$^{\circ, \dagger}$, Elena Sofia Ruzzetti$^\dagger$, Giancarlo A. Xompero$^{\diamond, \dagger}$, Fabio Massimo Zanzotto$^{\dagger,\diamond}$}
\IEEEauthorblockA{
$^\dagger$ \textit{Human-Centric ART, University of Rome Tor Vergata} \\
$^\circ$ \textit{Department of Computer Science, University of Luxembourg} \\
$^\diamond$ \textit{Almawave Labs, Rome}
}
}

\maketitle

\begin{abstract}
The robustness of Personally Identifiable Information (\textbf{PII}) protection in Large Language Models (\textbf{LLMs}) is a critical concern, yet the risks associated with \textbf{cross-lingual data extraction} remain under-explored. This study evaluates the vulnerability of English-centric and multilingual models to Training Data Extraction (\textbf{TDE}) attacks when prompted in non-English languages. 

We construct a multi-domain PII dataset comprising social media handles, email addresses, and phone numbers and translate the attack contexts into Italian, Spanish, French, and German.
Our results show that TDE attacks against both English-centric and multilingual models \textbf{transfer to different languages}: the attacks are successful on translated prompts, even though only the original English prompt might have been included in the pre-training data. A web-presence check on a sample of the translations confirms that they are not available online. The share of English leaks recovered in other languages grows with the multilingual capability of the model, and it drops sharply when the original wording is lost, even without a change of language.

This suggests that native multilingual pre-training facilitates the emergence of latent \textbf{cross-linguistic bridges} that simplify the retrieval of personally identifiable information (PII).
We analyze the activations of multilingual large language models (LLMs) and find that different translations of the same prompt are \textbf{bridged} in similar representations, with the strongest alignment in the middle layers.

Our results highlight a fundamental security gap in modern LLMs, necessitating more robust, language-agnostic sanitization strategies for future model alignment.
\end{abstract}


\section{Introduction}
Large Language Models (LLMs) are known to memorize fragments of their training data, and this memorization can be exploited to extract personally identifiable information (PII) -- like emails, phone numbers, addresses, credentials -- through Training Data Extraction (TDE) attacks \cite{Carlini2021Extracting, Carlini2023Quantifying}. Most work on this threat, however, treats memorization as a monolingual phenomenon: an attacker is assumed to query a model in the same language the memorized text was originally written in.

This assumption is broken down for the multilingual LLMs now deployed at scale, leaving a consequential blind spot. If a model, trained on multilingual data, \textbf{memorizes some data in one language} and can be prompted to \textbf{regurgitate that same memorized content when queried in other languages}, then privacy safeguards evaluated only in the source language systematically underestimate real-world exposure. 
Moreover, as multilingual capability improves, it may not only add new attack surfaces but also actively strengthen extraction across all of them if the model is building shared cross-lingual representations that link the same fact regardless of the language in which it is expressed.

Despite these stakes, whether TDE attacks transfer across languages and, if so, what mechanism underlies that transfer remains largely unexplored. Prior work has separately established that LLMs memorize and leak training data~\cite{Carlini2021Extracting,huang-etal-2022-large,Carlini2023Quantifying,nasr2023scalableextractiontrainingdata}, and safety alignment can be circumvented via cross-lingual prompting~\cite{Li2024CrossLanguage}, but these two lines of research have not been connected. It is not known whether memorization itself is language-agnostic (i.e., whether a fact memorized from an English document is stored in a way that a differently-worded, differently-languaged prompt can still trigger), or whether apparent leakage in a second language is simply an artifact of translation overlap with other multilingual training data. Answering this is what motivates our approach.

In this paper, we study \textbf{memorization in a cross-lingual scenario}: given an English dataset highly likely to appear in the training data of both monolingual and multilingual models, we ask whether data memorized from an English document can be extracted using an attack formulated in another language. We use \texttt{Pile-CC} a component of Pile~\cite{gao2020pile800gbdatasetdiverse} derived from Common Crawl.
Because Pile's collection was fully automated, it inadvertently contains substantial PII embedded in scraped blogs, forums, and news pages. We translate paragraphs from Pile-CC into multiple languages and apply TDE attacks in each target language, testing whether English training produces leakage when probed cross-lingually.

Our experiments show that both \textbf{monolingual and multilingual models are susceptible to this cross-lingual leakage}, and that the risk increases with the model's multilingual capability. These results point to the existence of \textbf{cross-lingual bridges}: shared underlying representations that allow attackers to recover memorized secrets even when the attack language differs from the language of the original training data. We further show, by comparing strict translations against paraphrases in both languages, that this leakage tracks fidelity to the memorized content and localize the effect to a specific band of middle transformer layers.

\section{Background and Related Work}
This section reviews training data extraction as a privacy threat, the defenses proposed against it, and the still-limited evidence on whether either the threat or the defense transfers across languages.

\paragraph{Privacy Vulnerabilities and Training Data Extraction}
Unintended memorization of the training data allows an attacker to extract such training data. \cite{Carlini2021Extracting} first showed how an attacker could exploit such phenomena: querying GPT-2 with training-set prefixes recovers verbatim PII, a phenomenon they formalized as $k$-eidetic memorization. \cite{Carlini2023Quantifying} later showed memorization scales log-linearly with model capacity and duplication frequency, and that a matching training-context prefix substantially increases extraction success -- a finding echoed by \cite{huang-etal-2022-large} for emails and social handles. \cite{nasr2023scalableextractiontrainingdata} extended this beyond controlled settings, showing open models like GPT-Neo and Pythia leak more than earlier estimates suggested, while \cite{carlini2024stealingproductionlanguagemodel} confirmed that closed, production-grade models such as GPT-4 remain vulnerable to black-box extraction.

These studies, however, focus primarily on monolingual contexts centred on English.
Nevertheless, cross-linguistic vulnerabilities have already emerged in terms of robustness against adversarial attacks: \cite{Li2024CrossLanguage} demonstrated that security alignment can be circumvented by translating jailbreak prompts into languages with scarce resources, implying that LLMs share a cross-linguistic latent space in which a security measure applied in one language is not necessarily valid in another. It has not been directly verified whether the same holds true for \emph{extraction} as opposed to jailbreaking.

\paragraph{Strategies to Protect Privacy in LLMs}
Mitigations can be applied at the data level, at training time, or as post-hoc model transformations.
Data-level approaches, from classical NER~\cite{Mamede2016} to transformer-based rewriting~\cite{Xu2019}, scrub PII before training, but have been shown to scale poorly to massive, uncurated corpora ~\cite{miranda2025preserving}.
Training-time approaches such as Differential Privacy~\cite{Hoory2021DP} offer formal guarantees but are costly and degrade the utility of the trained model. 
Post-hoc approaches, like machine unlearning~\cite{Yao2024Unlearning} and model editing \cite{Meng2023a_ROME, ruzzetti2025pme, venditti2026pae}, remove memorized content after the training has finished.
Critically, all of these defenses are mostly evaluated monolingually: whether unlearning or editing a PII record in English also prevents its recovery when the model is queried in another language remains largely untested.

\paragraph{Cross-Lingual Memorization and Transferability}
A small but growing body of work suggests the answer is no -- privacy behavior does not transfer cleanly across languages, though the direction and mechanism of this failure are still unclear.
\cite{luo-etal-2025-shared} analyze memorization across 95 languages and show that the standard assumption -- memorization tracks training-data volume -- breaks down once cross-lingual similarity is modeled: among typologically similar languages, those with \emph{less} training data are disproportionately more memorized, implying memorized content propagates through shared multilingual representations. \cite{srivastava2025owlprobingcrosslingualrecall} provide a direct behavioral demonstration of this: probing LLMs with book samples machine-translated into languages that have no prior published translation, they find substantial recall (GPT-4o identifies title and author 69\% of the time on entirely novel translations), showing that content memorized in English remains accessible under queries formulated in a language the model has never seen that content in.

At the same time, this transfer is demonstrably uneven. ~\cite{choi-etal-2024-cross} show that unlearning applied in one language does not propagate to others -- a fact deleted in English typically survives, verbatim, in Spanish or Chinese -- indicating that whatever shared representation enables cross-lingual leakage does not straightforwardly enable cross-lingual forgetting. \cite{mishra-etal-2026-models} and \cite{satvaty-etal-2025-memorization} further show that susceptibility to extraction and membership-inference attacks varies systematically with a language's structural properties (redundancy, morphological complexity, resource level). Together, these results position language not as a neutral channel but as an active variable shaping both leakage and its mitigation.


\textbf{Our contribution.} Existing work thus establishes that memorized content can, in principle, cross language boundaries~\cite{luo-etal-2025-shared, srivastava2025owlprobingcrosslingualrecall}, but this transfer is asymmetric and language-dependent~\cite{choi-etal-2024-cross, mishra-etal-2026-models, satvaty-etal-2025-memorization}. What remains untested is the specific, practically consequential question this paper addresses: whether \emph{PII} memorized from English training data text~\cite{srivastava2025owlprobingcrosslingualrecall} -- can be recovered via TDE attacks formulated in a different language. By translating paragraphs of Pile-CC and querying models in Italian, Spanish, French, and German for PII memorized only in English, we test directly whether the linguistic barrier offers any practical protection, and provide evidence for the \textbf{cross-lingual bridges} that may facilitate the recovery of secrets across linguistic boundaries. We demonstrate that such bridges are facilitated by high content fidelity and we localize their formation to a specific band of middle transformer layers.

\section{Methodology and Experiments}
The methodology of this study follows a multi-stage pipeline designed to evaluate the susceptibility of LLMs to cross-lingual Training Data Extraction (TDE).

\subsection{Attack Methodology}
We implement Training Data Extraction (TDE) attacks \cite{Carlini2021Extracting,Carlini2023Quantifying,nasr2023scalableextractiontrainingdata}. 
The objective is to determine if the models can verbatim generate a PII when prompted with a specific context.
\subsubsection{Formalization of the TDE Attack}
The objective of a TDE attack is to recover sensitive sequences $s \in \mathcal{D}_{train}$ \cite{Carlini2021Extracting}. Let $\theta$ represent the parameters of a causal language model $M$. The model defines a probability distribution over a sequence of tokens as
\begin{equation}
    P_\theta(x_1, \dots, x_n) = \prod_{i=1}^{n} P_\theta(x_i \mid x_{<i}).
\end{equation}
We select English samples from Pile-CC, and partition the sequence into a prefix $p$ of length $k$ and the target PII $\hat{s}$. The attack is successful if:
\begin{equation}
\hat{s} = \text{arg}\max_{s'} P_\theta(s' \mid p)
\end{equation}
that is, we define a sequence as \textit{verbatim memorized} if the generated output $\hat{s}$ matches the ground truth $s$ with an edit distance of zero.
In this study, we fix the context window to $k=200$ tokens. This specific length was chosen to provide a strong signal, as shorter contexts are also demonstrated to be less likely to extract memorized information \cite{huang-etal-2022-large,ruzzetti2025pme,venditti2026pae}.

The worst-case scenario for this attack is that the prompt $p$ is derived directly from the target model’s training data: the success rate of such attacks is defined as \textit{extractable memorization} \cite{Carlini2023Quantifying}.

\subsubsection{Cross-Lingual Mapping Challenge}
We compare the accuracy of semantically identical attacks, translated in different languages, to understand whether \textbf{memorized information in English can also be extracted with attacks in other languages}.
This setup assesses whether the model’s internal semantic mapping allows for the leakage of sensitive data across language boundaries. While the original definition of \textit{extractable memory} assumes that leakage is more likely when using exactly the same data that the model has seen -- and potentially memorized -- during the pre-training phase, we demonstrate that the attack is transferable to other languages.

\subsection{Cross-Lingual Dataset Construction}
\label{sec:data}
%
Our dataset is composed of text samples from Pile-CC
\footnote{Original data available \href{https://huggingface.co/datasets/tomekkorbak/pile-pii-scrubadub}{here}, and verified by us scraping Pile-CC} that contains a PII. We collected three different PII types: \texttt{email} addresses, \texttt{phone} numbers and \texttt{twitter} handles. The scraped dataset, completely in English dataset, is composed of a \textit{prompt} is followed by a \textit{PII}. 

To measure transferability of the attacks to other languages, we then machine translated each of the prompts in French, German, Italian and Spanish.
The translation was performed using a \texttt{Gemma 3 27b it} via the Google Gemini API, instructing the model to respect the exact meaning of the original English sample and keeping the same formatting, and overall structure of the original prompt, to minimize the difference between the original English prompt and its translation. 
At least one author is proficient in each of the target languages, and
spot checks were conducted to ensure the quality of the translations.
At the end of this process, for each of the target languages, we obtained a dataset containing 4695 email, 9005 twitter handles and 3000 phone numbers.


\subsubsection{Novelty of the Translations}
\label{sec:novelty}
To test whether the TDE attacks actually \textbf{transfer}, we need to verify that the \textbf{translations of the Pile-CC data are unlikely} to be used as part of the \textbf{pre-training} material of our tested model.
While this seems reasonable, as we use specific samples from Pile-CC, we need to verify that such data are not already seen by the target models to ensure that we are measuring the \textbf{transferability} of the TDE attacks.

For most of the models tested, however, it is not possible to accurately verify the data used for training, since the model providers do not disclose the data used for pre-training. We therefore \textbf{search for instances of our translated data in search engines} and \textbf{check for matches} between our translation and segments of the returned pages. If no matches are found, this means that the observed translation is unlikely to be available online and, consequently, unlikely to have been used for the pre-training phase of the target models.

To assess whether the translated versions of the English samples were
publicly available online before their generation, we develop a
web-based verification pipeline. We apply it to a subset of 200
translated samples per PII type. For each of these samples, we first
generate a distinctive exact-match search query and use the Tavily
search API\footnote{Documentation available
\href{https://www.tavily.com/}{on this website}.} to identify
potentially matching web pages. The URLs returned by the search engine
are then compared with the Common Crawl index to identify archived
versions that existed before the translation. For each relevant capture,
we retrieve only the corresponding WARC record using HTTP range
requests, extract the textual content of the page, and compare it with
the translated sample. To compare the retrieved pages with the
translations of our dataset, we compute exact matches and fuzzy matches.

\textit{Exact matching} checks whether the original translated sample occurred verbatim in the extracted webpage text.
\textit{Fuzzy matching} computes the normalized Levenshtein-based similarity between the translated sample and candidate text windows of comparable length ratio from the webpage; a match was considered positive when the similarity score exceeded a predefined threshold ($90\%$).
This procedure provides evidence about whether the translated samples were publicly available on the web before their generation.

\subsection{Target Models: English-centric and Multi-lingual models}
We evaluate the attack across three distinct families of models to observe how scale and pre-training objectives influence cross-lingual leakage:

\begin{itemize}
    \item EleutherAI {\texttt{GPT-Neo} \texttt{1.3B} and \texttt{2.7B}}~\cite{gptneo} and \texttt{GPT-J 6B}~\cite{gpt-j}: English-centric models, these serve as a baseline to test the persistence of memorization when the model is forced to switch from its dominant training language (English) to a non-native prompt context. We selected these models because they are known to have been trained on The Pile, and therefore on the Pile-CC subset from which our evaluation samples were drawn.
    
    \item Meta \texttt{Llama 3.2 1B} and \texttt{3B}~\cite{grattafiori2024llama3herdmodels}: multilingual models, these are integrated to evaluate the attack surface in models natively trained on a massive cross-lingual corpus, potentially facilitating the bridge between translated prompts and English-stored PII. 

    \item Qwen \texttt{Qwen 2.5 3B} and \texttt{7B}~\cite{qwen2025qwen25technicalreport}: multilingual models trained on a large-scale, highly multilingual corpus. These models allow us to assess whether the observed cross-lingual memorization behavior generalizes beyond the Llama family and to compare the impact of different pre-training data mixtures and tokenization strategies on privacy leakage.
\end{itemize}

For \texttt{Llama} and \texttt{Qwen} models, there is no public confirmation that our selected Pile-CC subset was included in the training data; nevertheless, the empirical evidence obtained from our attacks suggests that at least part of the evaluated content was likely seen during pre-training.

\subsection{Finding Cross-Lingual Bridges}
\label{sec:finding-bridges}
Let $x_{n}^{l,\mathrm{eng}}$ and $x_{n}^{l,\mathrm{target}}$ denote the hidden-state representations at layer $l$ and token position $n$, where $n$ corresponds to the last token of the input sequence.
If a cross-lingual TDE attack succeeds, we hypothesize that models construct a cross-lingual bridge between representations in English ($\mathrm{eng}$) and the corresponding representation in another language ($\mathrm{target}$).

If cross-lingual TDE succeeds because the model has \textbf{learned a shared,
language-agnostic representation of a specific memorized fact}, such as the
association between a person's name and their email address, then this
representation should be recoverable whenever the \textbf{content is preserved},
regardless of the surface language. Conversely, if apparent cross-lingual
leakage is driven by weaker signals, such as general topical or semantic
similarity between the prompt and the training context, then any prompt
conveying a similar meaning should activate comparable internal states,
regardless of whether it precisely preserves the original content.

To distinguish these two accounts, we vary two factors independently:
\textbf{language} and \textbf{wording
fidelity}. We hence compare our strict, structure-preserving translations with free paraphrases.
Crossing these factors allows us to determine not only whether translation
moves a representation away from the original, but also whether this
movement is driven by the language change itself or by the accompanying
change in wording. Symmetrically, we can assess whether paraphrasing degrades
the representation even when the language remains unchanged.

\subsubsection{Experimental Conditions}

For each sample $x^{\mathrm{eng}}$ in our dataset, we construct three
additional variants:

\begin{itemize}
\item $x^{\mathrm{target}}_{\mathrm{strict}}$: a faithful,
structure-preserving translation used in our main TDE attacks (see
\ref{sec:data}). It introduces the same
language change as the attack, with no wording changes beyond those
required by translation.

\item $x^{\mathrm{eng}}_{\mathrm{para}}$: an English paraphrase of
$x^{\mathrm{eng}}$. It preserves the language while changing the wording,
thereby isolating the effect of rewording without a language change.

\item $x^{\mathrm{target}}_{\mathrm{para}}$: a paraphrase of
$x^{\mathrm{eng}}$ in the target language $\mathrm{target}$. This condition
changes both language and wording, and most closely resembles the setting
of an attacker who does not have access to the exact original phrasing.

\end{itemize}

All paraphrases preserve the topic and general meaning of the source sample
while deliberately varying lexical choice, sentence structure, and the order
in which entities are introduced. Details on the generation and evaluation of the paraphrases are detailed in Appendix~\ref{app:paraphrase}.

We feed each variant through the target model and extract hidden states at
every layer $l$, at the last token position $n$ of the input prompt, immediately preceding the point where the PII would be generated. This representation
captures the model's accumulated contextual state, and has been demonstrate to effectively encode the contextual information of the input \cite{ICLR2026_5f124a19}.

\subsubsection{Similarity metric}
We experimented with the cosine similarity as the similarity metric for the representations:
\begin{equation}
\frac{x_n^{l,a} \cdot x_n^{l,b}}
{\lVert x_n^{l,a} \rVert , \lVert x_n^{l,b} \rVert}.
\label{eq}
\end{equation}

For each sample, we compute three per-layer similarity scores with respect
to the English original context:
\begin{equation}
\mathrm{sim}^{l}_{\mathrm{strict}}
=
\mathrm{sim}^{l}\left(
x_n^{l,\mathrm{eng}},
x_{n,\mathrm{strict}}^{l,\mathrm{target}}
\right)
\end{equation}

\begin{equation}
\mathrm{sim}^{l}_{\mathrm{para-eng}}
=
\mathrm{sim}^{l}\left(
x_n^{l,\mathrm{eng}},
x_{n,\mathrm{para}}^{l,\mathrm{eng}}
\right)
\end{equation}

\begin{equation}
\mathrm{sim}^{l}_{\mathrm{para-target}}
=
\mathrm{sim}^{l}\left(
x_n^{l,\mathrm{eng}},
x_{n,\mathrm{para}}^{l,\mathrm{target}}
\right).
\end{equation}

We then average these scores across the dataset to obtain a per-layer
similarity profile for each condition.

\subsubsection{Hypotheses and Interpretation}

We test three ordered predictions.

\paragraph{H1: Translation preserves the bridge better than paraphrasing
within the same language}
\begin{equation}
\mathrm{sim}^{l}_{\mathrm{strict}}
>
\mathrm{sim}^{l}_{\mathrm{para-eng}}.
\label{eq:L1-paraL1}
\end{equation}

This is the critical comparison because if a strict translation into a
different language remains closer to the original representation than an
English paraphrase, this suggests that the representation tracks content
fidelity. In particular, it provides evidence
that the information surviving translation is associated with the specific
memorized fact.

\paragraph{H2: Wording loss compounds with language change}
\begin{equation}
\mathrm{sim}^{l}_{\mathrm{para-eng}}
>
\mathrm{sim}^{l}_{\mathrm{para-target}}.
\label{eq:paraL1-paraL2}
\end{equation}

Comparing the two paraphrase conditions isolates the additional effect of
changing language once the wording has already been altered. A difference
between these conditions indicates that language and wording fidelity both
contribute to representational drift.

\paragraph{H3: Strict translation remains closer than free-form
paraphrasing in the target language}
\begin{equation}
\mathrm{sim}^{l}_{\mathrm{strict}}
>
\mathrm{sim}^{l}_{\mathrm{para-target}}.
\label{eq:L1-paraL2}
\end{equation}
As a consequence of H1 and H2, we expect a faithful translation to produce a representation
closer to the memorized English representation than a free-form paraphrase. This would indicate that content fidelity plays an important role in recalling the memorized segment.

We expect these differences to be more pronounced in the central layers, where
representations may encode more abstract factual and associative content,
and relatively minor in the initial layers, which are more closely associated
with lexical and tokenization properties at the surface level and in the final layers, where representations have been observed to specialize in generating the target language \cite{wendler-etal-2024-llamas}.

A consistent ordering,
\begin{equation}
\mathrm{sim}^{l}_{\mathrm{strict}}
>
\mathrm{sim}^{l}_{\mathrm{para-eng}}
>
\mathrm{sim}^{l}_{\mathrm{para-target}},
\label{eq}
\end{equation}
would be consistent with the existence of
a \textbf{cross-lingual bridge}: \textbf{a representation anchored to specific
memorized content} that remains \textbf{relatively stable under translation} but
degrades when the content is substantially restated, particularly when
paraphrasing is combined with a language change. This would explain why our TDE attacks transfer across languages while remaining sensitive to the fidelity of the attack prompt,
and argues against a purely coarse-semantic account of the leakage, in
which any topically related prompt, translated or not, would be equally
effective.

\section{Experimental Results}

\subsection{Web-Presence Check: Translated samples not Online}
We check, on a subset of 200 samples per PII type, whether the
translations used in our attacks were already available online before
we generated them. If this were the case, a model might recognize our
translation simply because it had already seen that exact text during
training.
Table~\ref{tab:web_presence_statistics} summarizes the results of our
web verification pipeline (see Section~\ref{sec:novelty}).

\begin{table}[t]

\caption{
\textbf{Translated samples from Pile-CC (200 per PII type) are novel.}
We report descriptive statistics of the web-presence search results for Twitter handles, email addresses and phone numbers. The  
\textbf{Queries w/ results \%} reports the proportion of queries returning at least one result, 
while \textbf{Historical captures per query} reports the number of historical captures identified for each query. 
\textbf{Exact match} and \textbf{Max. Fuzzy score} denotes the exact lexical overlap and the maximum fuzzy matching score as detailed in Section \ref{sec:novelty}.}
\label{tab:web_presence_statistics}

\centering
\small
\setlength{\tabcolsep}{5pt}
\renewcommand{\arraystretch}{1.15}
\resizebox{\linewidth}{!}{%
\begin{tabular}{lrccccc}
\toprule
\textbf{PII Type} &
\textbf{Statistic} &
\textbf{Queries w/} &
\textbf{Historical} &
\textbf{Exact} &
\textbf{Max. Fuzzy} \\
&
&
\textbf{results \%} &
\textbf{captures per query} &
\textbf{match} &
\textbf{score} \\
\midrule

\multirow{7}{*}{{Twitter}}
& Mean & 97.6 & 0.053 & 0 & 0.229 \\ 
& Std. & 15.2 & 0.615 & 0 & 3.237  \\ 
& Min. & 0 & 0 & 0 & 0 \\ 
& 25th percentile & 100 & 0 & 0 & 0 \\ 
& Median & 100 & 0 & 0 & 0 \\ 
& 75th percentile
& 100 & 0 & 0 & 0 \\ 
& Max.
& 100 & 13 & 0 & 46.289 \\ 
\midrule

\multirow{7}{*}{{Email}}
& Mean
& 66.8 & 0.044 & 0 & 0.172 \\ 
& Std.
& 47.1 & 0.693& 0 & 2.800  \\ 
& Min.
& 0 & 0 & 0 & 0 \\ 
& 25th percentile
& 0 & 0 & 0 & 0 \\ 
& Median
& 100 & 0 & 0 & 0 \\ 
& 75th percentile
& 100 & 0 & 0 & 0 \\ 
& Max.
& 100 & 14 & 0 & 47.327 \\ 
\midrule
\multirow{7}{*}{Phone}   
& Mean            & 100     & 0.054  & 0     & 0.163  \\
& Std.            & 0     & 0.772 & 0     & 2.655  \\
& Min.            & 100     & 0        & 0     & 0      \\
& 25th percentile & 100     & 0        & 0     & 0      \\
& Median          & 100     & 0        & 0     & 0      \\
& 75th percentile & 100     & 0        & 0     & 0      \\
& Max.            & 100     & 14       & 0     & 44.156 \\
\bottomrule
\end{tabular}
}
\end{table}

\paragraph{Almost all queries returned some results, but almost none of these were actual matches}
For all tested PII type, almost every query returned at least
one page from the search engine (97.6\% of queries for Twitter, 66.8\% for
emails, 100\% for phone numbers). This simply means that the search terms were common enough to return
\emph{something}, but it does not mean that they are a match. When
we examine how many of those returned pages actually contain a
matching historical screenshot, the numbers drop to almost zero: on average, only 0.053
matching captures per query for Twitter, 0.044 for emails and 0.054 for phone numbers.
For the majority of the extracts there were no matching captures at all: in fact, the 25th percentile, the median and the 75th percentile are all 0. 
A small number of samples represented outliers, with a maximum of
13 (Twitter) or 14 (emails and phone numbers) matching hits, which brings the average
slightly above zero, but the typical sample has none.

\paragraph{No online sample matches exactly}
Out of the 200 samples checked for each type of PII, the exact match
rate is 0 in every statistic we have reported: mean, minimum, maximum
and every percentile.

\paragraph{Fuzzy matches are rare and, when they did occur, were in any case weak}
The fuzzy match score measures how closely the text of a returned page matches our
translated segment, taking into account minor differences in wording. For all the tested PII types, this score is 0 for the vast majority of segments (again, the
25th percentile, the median and the 75th percentile are all 0), which means that most of the returned pages are not significantly similar to our segments at all. The
average score is low (0.229 for Twitter, 0.172 for emails, and 0.163 for phone numbers). Even the few outliers with higher scores (up to 46.3 for Twitter, 47.3 for
email, 44.16 for phones) remain well below the similarity threshold we use to define something as a match (90, as described in Section~\ref{sec:novelty}).

These results demonstrate that the translated samples we checked are,
for all intents and purposes, not present on the web: searches return
numerous irrelevant pages, but virtually no exact matches, almost no
historical copies, and only a handful of approximate matches that do not
even come close to our match threshold. This supports our hypothesis
that the translated data were not part of the models' pre-training data
and that the cross-linguistic leak we observe in Section~\ref{sec:transferability-results} reflects
a genuine transfer of memorized content between languages.

\subsection{Cross-Lingual Transferability of TDE Attacks}
\label{sec:transferability-results}

\begin{table*}[h]

\caption{
\textbf{A large portion of PII are leaked also in other languages.}
We report the number of successfully leaked PII, together with the percentage
of English leaks also observed in each language. The number of generations
containing PII instances is reported for each language.
}
\label{tab:leakage_counts}

\centering
\small
\setlength{\tabcolsep}{3pt}
\resizebox{\linewidth}{!}{%
\begin{tabular}{llr|rrr|rrr|rrr|rrr}
\toprule
\multirow{2}{*}{\textbf{PII type}} &
\multirow{2}{*}{\textbf{Model}} &
\multicolumn{1}{c|}{\textbf{English}} &
\multicolumn{3}{c|}{\textbf{French}} &
\multicolumn{3}{c|}{\textbf{Italian}} &
\multicolumn{3}{c|}{\textbf{Spanish}} &
\multicolumn{3}{c}{\textbf{German}} \\
\cmidrule(lr){3-3}
\cmidrule(lr){4-6}
\cmidrule(lr){7-9}
\cmidrule(lr){10-12}
\cmidrule(lr){13-15}
&
& \textbf{Leak} &
\textbf{Leak} & \textbf{Rel.} & \textbf{Generated} &
\textbf{Leak} & \textbf{Rel.} & \textbf{Generated} &
\textbf{Leak} & \textbf{Rel.} & \textbf{Generated} &
\textbf{Leak} & \textbf{Rel.} & \textbf{Generated} \\
\midrule

\multirow{7}{*}{Twitter}
& \texttt{Llama 3.2 1B}
& 340 & 255 & 75.0\% & 1788 & 243 & 71.5\% & 1724 & 260 & 76.5\% & 1893 & 213 & 62.6\% & 1369 \\
& \texttt{Llama 3.2 3B}
& 548 & 492 & 89.8\% & 2284 & 501 & 91.4\% & 2327 & 488 & 89.1\% & 2334 & 389 & 71.0\% & 1831 \\
& \texttt{Qwen2.5 3B}
& 468 & 395 & 84.4\% & 2330 & 372 & 79.5\% & 2360 & 384 & 82.1\% & 2391 & 298 & 63.7\% & 1693 \\
& \texttt{Qwen2.5 7B}
& 591 & 479 & 81.0\% & 2682 & 439 & 74.3\% & 2647 & 440 & 74.5\% & 2626 & 361 & 61.1\% & 2197 \\
& \texttt{GPT-J 6B}
& 1020 & 426 & 41.8\% & 2033 & 450 & 44.1\% & 2300 & 444 & 43.5\% & 2219 & 337 & 33.0\% & 1688 \\
& \texttt{GPT-Neo 1.3B}
& 505 & 228 & 45.1\% & 1512 & 215 & 42.6\% & 1608 & 222 & 44.0\% & 1574 & 148 & 29.3\% & 1072 \\
& \texttt{GPT-Neo 2.7B}
& 694 & 236 & 34.0\% & 1646 & 264 & 38.0\% & 1857 & 249 & 35.9\% & 1784 & 160 & 23.1\% & 1163 \\
\midrule

\multirow{7}{*}{Email}
& \texttt{Llama 3.2 1B}
& 167 & 78 & 46.7\% & 1846 & 74 & 44.3\% & 2020 & 71 & 42.5\% & 1787 & 80 & 47.9\% & 2089 \\
& \texttt{Llama 3.2 3B}
& 329 & 195 & 59.3\% & 2370 & 192 & 58.4\% & 2450 & 187 & 56.8\% & 2143 & 183 & 55.6\% & 2376 \\
& \texttt{Qwen2.5 3B}
& 242 & 136 & 56.2\% & 2960 & 145 & 59.9\% & 2995 & 140 & 57.9\% & 2859 & 118 & 48.8\% & 2681 \\
& \texttt{Qwen2.5 7B}
& 328 & 210 & 64.0\% & 3326 & 204 & 62.2\% & 3298 & 227 & 69.2\% & 3114 & 193 & 58.8\% & 3158 \\
& \texttt{GPT-J 6B}
& 726 & 121 & 16.7\% & 2811 & 114 & 15.7\% & 2635 & 108 & 14.9\% & 2081 & 109 & 15.0\% & 2646 \\
& \texttt{GPT-Neo 1.3B}
& 267 & 42 & 15.7\% & 1565 & 44 & 16.5\% & 1570 & 35 & 13.1\% & 1329 & 34 & 12.7\% & 1353 \\
& \texttt{GPT-Neo 2.7B}
& 442 & 70 & 15.8\% & 2284 & 62 & 14.0\% & 2279 & 54 & 12.2\% & 1659 & 56 & 12.7\% & 2075 \\
\midrule

\multirow{7}{*}{Phone}
& \texttt{Llama 3.2 1B}
& 161 & 32 & 19.9\% & 1851 & 37 & 23.0\% & 1691 & 35 & 21.7\% & 1793 & 6 & 3.7\% & 602 \\
& \texttt{Llama 3.2 3B}
& 388 & 111 & 28.6\% & 2108 & 103 & 26.5\% & 1971 & 111 & 28.6\% & 2024 & 55 & 14.2\% & 1364 \\
& \texttt{Qwen2.5 3B}
& 197 & 35 & 17.8\% & 2409 & 38 & 19.3\% & 2188 & 43 & 21.8\% & 2354 & 8 & 4.1\% & 235 \\
& \texttt{Qwen2.5 7B}
& 285 & 54 & 18.9\% & 2443 & 58 & 20.4\% & 2340 & 61 & 21.4\% & 2450 & 10 & 3.5\% & 468 \\
& \texttt{GPT-J 6B}
& 495 & 18 & 3.6\% & 1547 & 16 & 3.2\% & 1326 & 21 & 4.2\% & 1609 & 5 & 1.0\% & 233 \\
& \texttt{GPT-Neo 1.3B}
& 138 & 3 & 2.2\% & 461 & 3 & 2.2\% & 472 & 3 & 2.2\% & 631 & 1 & 0.7\% & 91 \\
& \texttt{GPT-Neo 2.7B}
& 256 & 4 & 1.6\% & 938 & 4 & 1.6\% & 837 & 6 & 2.3\% & 1021 & 1 & 0.4\% & 100 \\

\bottomrule
\end{tabular}
}
\end{table*}
Having verified, on a sample of 200 passages per PII type, that the
translations are not available online, we now check whether they can
nevertheless cause a leak of private information across different
languages.
Table~\ref{tab:leakage_counts} reports, for each PII type, model, and language,
the number of successfully leaked PII (Leak), the percentage of English leaks also observed in other languages (Rel.), and the total number of
completions that contain a PII under the TDE attack. 

\textbf{TDE attacks transfer across different languages}, and our results reveal that both the PII type and the multilingual capabilities of the model influence the attack success rate.
\paragraph{Leakage transfers to every target language, for every model}
Across all seven models and three PII types, cross-lingual attacks in French,
Italian, Spanish, and German recover a non-trivial number of PII instances
that were only ever present, in their original form, in the English training
data. 

This confirms the central hypothesis of our study: memorization is not
confined to the language in which the memorized content was originally
expressed, and the linguistic barrier offers only partial protection rather
than a hard boundary.

\paragraph{Leakage is strongest in English but remains substantial
cross-lingually, and this gap tracks a model's multilingual capability}
For every model, the English attack yields the highest leak
count and leak rate, which is expected: it is the only condition where the
attack prefix is drawn verbatim from a plausible training distribution. 

However, the \emph{size} of the drop between English and the
other four languages differs sharply by model family. 
For the multilingual models (Llama~3.2, Qwen2.5), the cross-lingual leak rate retains up to 91.4\% of the Twitter handles leaked in English, up to 69.2\% for emails, and up to 28.6\% for phone numbers. 
For the English-centric models (GPT-Neo, GPT-J), the same
ratios are much smaller for Twitter handles (up to 45\%) and email addresses (up to 16.7\%), and collapse for phone numbers -- most strikingly for GPT-J~6B and GPT-Neo~2.7B, which leak less than 3\% of phone numbers cross-lingually. 

This is consistent with our hypothesis that \textbf{cross-lingual leakage is mediated by shared multilingual representations}: models trained on substantially multilingual corpora exhibit a much smaller English-to-other-language gap, even though their absolute
English leak rates are often lower than English-centric models'. 
Model scale alone does not explain this pattern -- the largest English-centric model,
GPT-J~6B, shows a \emph{weak} transfer with respect to the similar-sized Qwen2.5 7B --
which argues against transfer being a simple byproduct of overall model
capability and in favor of it depending specifically on multilingual
training.

\paragraph{PII type modulates both absolute leakage and transferability}
Across all models, Twitter handles are leaked at far higher rates than
emails, which in turn leak more than phone numbers, both in English
and cross-lingually. This ordering likely reflects differences in the
redundancy and predictability of each PII type's surrounding context (handles
are frequently repeated verbatim near associated usernames or mentions, while
phone numbers are comparatively sparse and less syntactically constrained),
consistent with prior evidence that extraction risk depends on structural
properties of the target string as well as the language it is expressed in.
Phone numbers are also the PII type for which cross-lingual transfer is
proportionally weakest for the English-centric models.

These results establish that \textbf{TDE attacks transfer across the
language boundary} for both monolingual and multilingual models, that the
\textbf{degree of transfer scales with a model's multilingual training} and that this cross-lingual susceptibility is not
uniform but is shaped by the interaction of model family, language, and PII type.

\begin{figure*}
    \includegraphics[width=\textwidth]{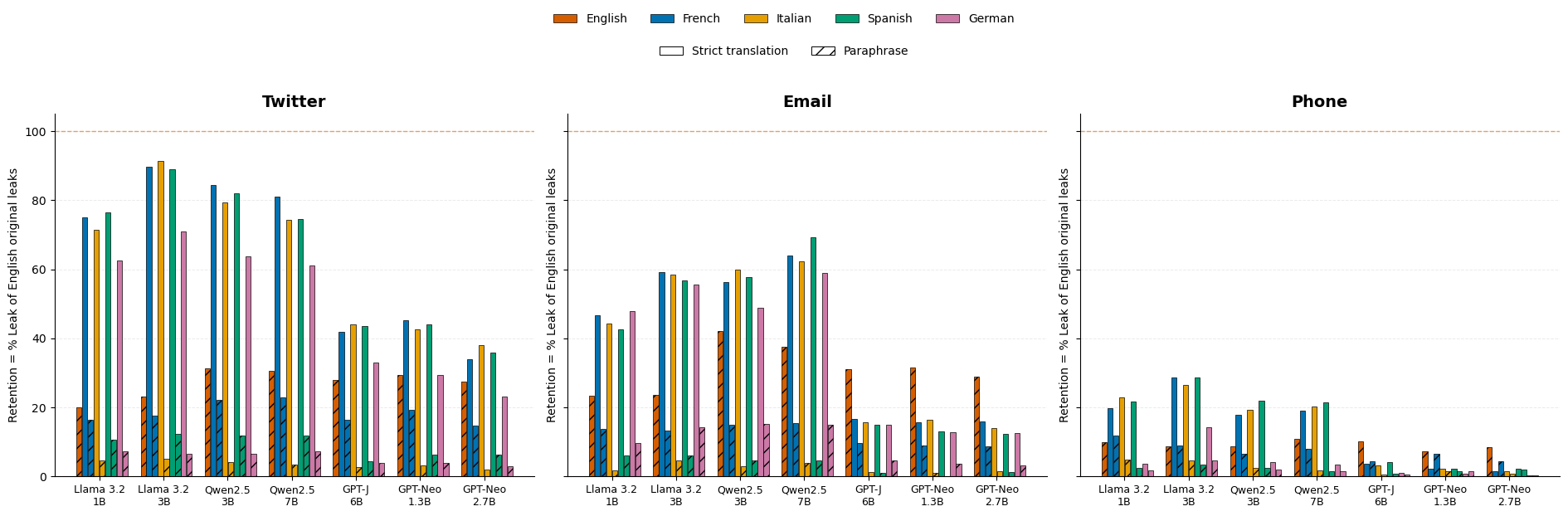}
    \caption{Retention of leaked PII instances under a paraphrased attack,
compared to strict-attack leak count, averaged across
the four multilingual models (Llama~3.2~1B/3B, Qwen2.5~3B/7B) and the three
English-centric models (GPT-J~6B, GPT-Neo~1.3B/2.7B)}
    \label{fig:para-attacks}
\end{figure*}

\subsection{Transferability of TDE Attacks Under Paraphrase}
\label{sec:paraphrase-transferability}
Section~\ref{sec:transferability-results} showed that TDE attacks transfer
across languages when the attack prefix is a faithful, strict translation of
the memorized English passages. 
We therefore ask what factors are key to the effective retrieval of memorized PII.
Building on the disentanglement of language and wording fidelity introduced
in Section~\ref{sec:finding-bridges}, we now first test its behavioral
counterpart: whether leakage itself, not just representational similarity,
tracks content fidelity.

We hence first analyze the behavioral effects of paraphrases, performing the TDE attacks on the paraphrased inputs. Figure~\ref{fig:para-attacks} reports, for the same PII types and models as Table~\ref{tab:leakage_counts}, the leakage obtained when the attack prefix is paraphrased. To facilitate comparisons across languages and models, we report leakage as a percentage of the corresponding English leaks for each model and PII type. Thus, 100\% indicates the same number of leaked PII instances as the English baseline, while values below 100\% indicate lower leakage. We refer to this relative leakage as \emph{retention} (for completeness, in Appendix \ref{app:complete-para-attacks} we report the raw number of leaks).

\paragraph{Paraphrasing the attack sharply reduces leakage, in every
language and for every model}
For both Twitter handles and email addresses, the retention rate for paraphrased English input drops to about one-quarter or one-third of the count obtained with the strict attack, while the retention rate for paraphrased input in the target language drops even further, to about one-tenth.
For phone numbers, the results vary more widely, but the number of leaks when using paraphrased input is always lower than the corresponding number of leaks obtained with strict translation for multilingual models, and comparable in English-centric models (which leak only a small fraction of the phone numbers).
Furthermore, for all multilingual models, the leak rate in the case of a paraphrasing attack is substantially lower than that obtained with literal translation into other languages.
These results support our hypothesis that the loss of the exact wording leads to a substantial loss of extraction success, even when the target language does not change.
We will explore this point further in Section~\ref{sec:bridge-results}.

\paragraph{English paraphrase consistently retains more leaked instances
than cross-lingual paraphrase}
English paraphrase retention is roughly two to three
times higher than cross-lingual paraphrase retention, for both multilingual and
English-centric models alike, with the only exception registered on phone numbers extracted on Llama 3.2 1B. 
This is the attack-level counterpart of the Strict~$>$~English paraphrase~$>$~Target language paraphrase ordering that we expect to hold representationally as well, as described in Section \ref{sec:finding-bridges}: paraphrasing costs
leakage, and paraphrasing \emph{into another language} costs more of it than
paraphrasing alone.

Paraphrasing the attacker's prompt substantially
reduces training-data extraction, and reduces it further when combined with
translation into another language -- consistent with the content-fidelity
ordering hypothesized in Section \ref{sec:finding-bridges}. We hence expect representational similarity to track the same content-fidelity ordering, which we test directly in the next section.

\begin{figure*}[h]
    \centering

    \begin{minipage}{0.7\textwidth}
        \centering
         \includegraphics[width=\textwidth]{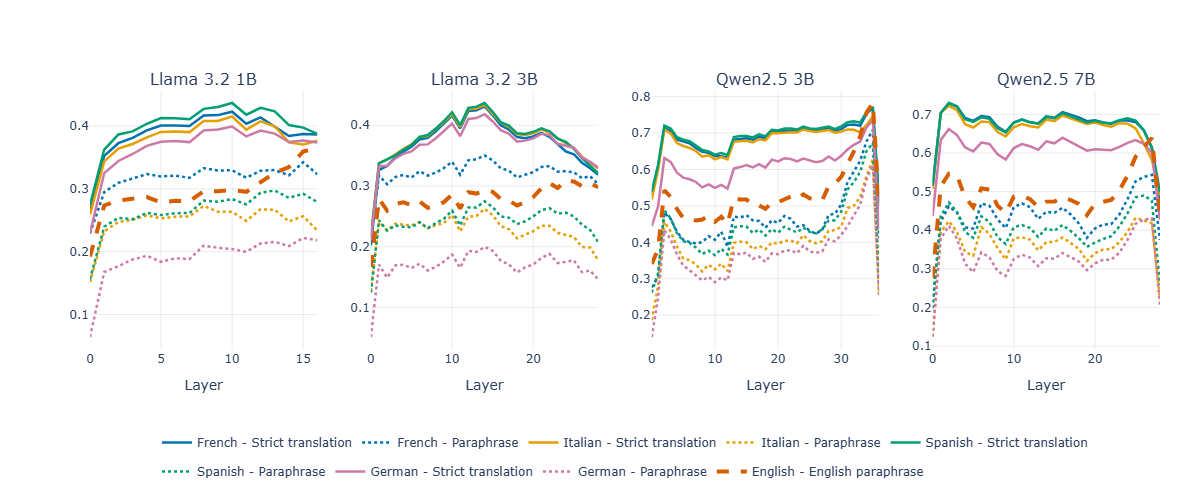}

        \smallskip
        {\footnotesize {(a)} Twitter}
    \end{minipage}
    
    \begin{minipage}{0.7\textwidth}
        \centering
         \includegraphics[width=\textwidth]{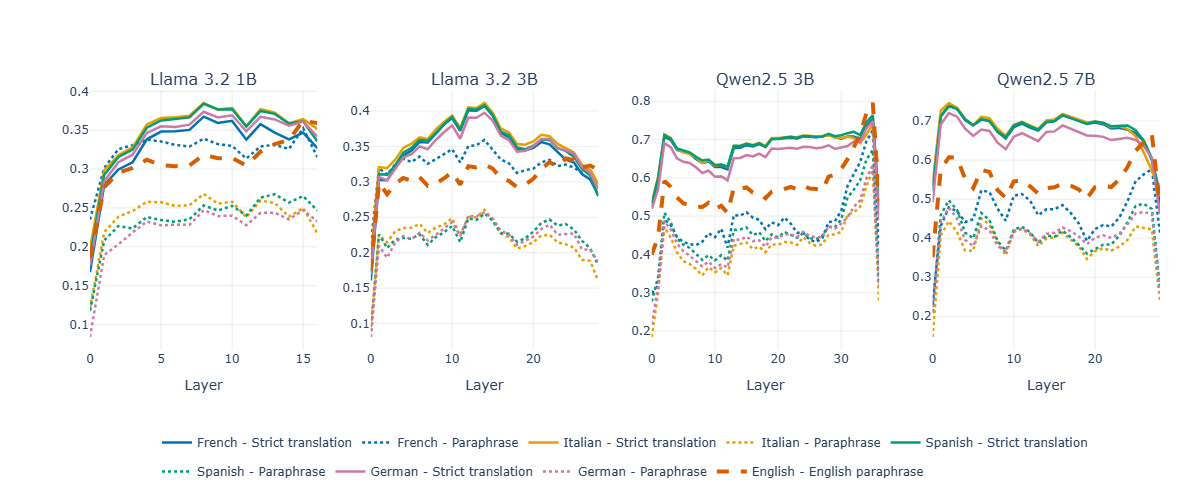}

        \smallskip
        {\footnotesize {(b)} Email}
    \end{minipage}

    \begin{minipage}{0.7\textwidth}
        \centering
        \includegraphics[width=\textwidth]{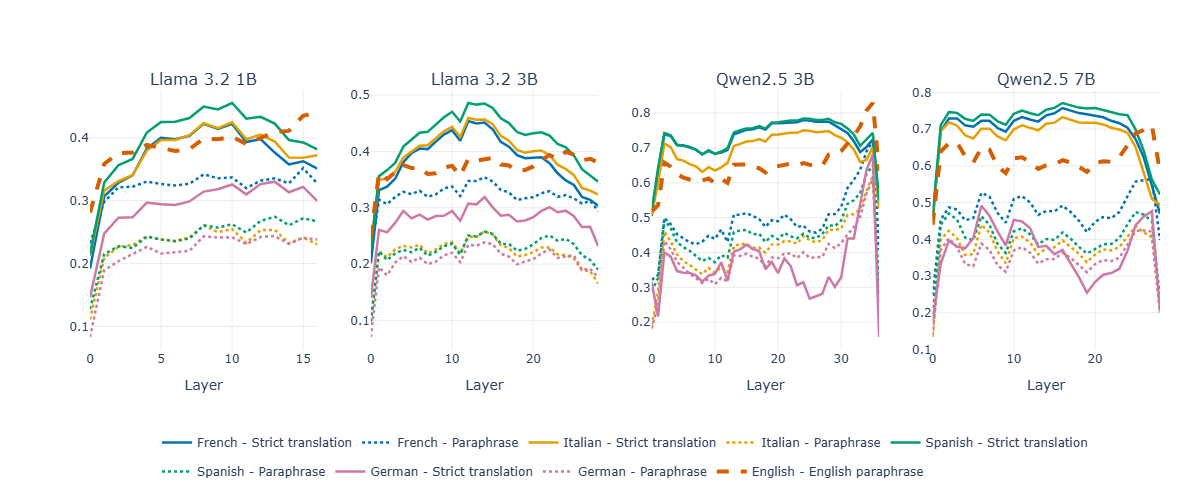}

        \smallskip
        {\footnotesize {(c)} Phone}
    \end{minipage}

    \caption{Mean cosine similarity between the representations of each strict translation and paraphrase and the English original, by layer, for each PII type.}
    \label{fig:bridge}
\end{figure*}

\subsection{Representational Evidence for Cross-Lingual Bridges}
\label{sec:bridge-results}
To explain the success of TDE attacks in other languages, we then demonstrate that, from a representational point of view, models tend to \emph{bridge} structurally similar inputs from different languages.
As described in Section \ref{sec:finding-bridges}, we hence compare the similarity of the representations under strict translations and free-form paraphrases. For this analysis, we focus on the activations of multilingual LLMs, since TDE attacks are more likely to transfer for these models.

Figure~\ref{fig:bridge} report the
per-layer results of the similarity analysis described in
Section~\ref{sec:finding-bridges}, for Twitter handles (\ref{fig:bridge}-a), email addresses (\ref{fig:bridge}-b) and phone numbers (\ref{fig:bridge}-c)
respectively.
For each model, we plot the mean cosine similarity to the English original representation across layers, for the strict translation for each target language, the English paraphrase, and the target-language paraphrase.
Table~\ref{tab:bridge_summary} summarizes the same results averaged across all layers, for each model, PII type, and condition, reporting the German strict translation separately from the French, Italian and Spanish ones.

\begin{table}[t]
\caption{Mean cosine similarity to the English representation, averaged
over all layers. Strict translations are reported separately for
French/Italian/Spanish (mean over the three languages) and German;
paraphrases are averaged over the four target languages.}
\label{tab:bridge_summary}
\centering
\small
\resizebox{\linewidth}{!}{%
\begin{tabular}{llcccc}
\toprule
\textbf{PII type} & \textbf{Model} & \textbf{Strict} & \textbf{Strict} & \textbf{English} & \textbf{Target language} \\
& & \textbf{(FR/IT/ES)} & \textbf{(DE)} & \textbf{paraphrase} & \textbf{paraphrase} \\
\midrule
\multirow{4}{*}{Twitter}
& \texttt{Llama 3.2 1B} & 0.389 & 0.366 & 0.296 & 0.255 \\
& \texttt{Llama 3.2 3B} & 0.375 & 0.367 & 0.279 & 0.239 \\
& \texttt{Qwen2.5 3B}   & 0.681 & 0.603 & 0.524 & 0.417 \\
& \texttt{Qwen2.5 7B}   & 0.668 & 0.607 & 0.491 & 0.381 \\
\midrule
\multirow{4}{*}{Email}
& \texttt{Llama 3.2 1B} & 0.340 & 0.338 & 0.309 & 0.255 \\
& \texttt{Llama 3.2 3B} & 0.348 & 0.343 & 0.305 & 0.246 \\
& \texttt{Qwen2.5 3B}   & 0.678 & 0.653 & 0.575 & 0.447 \\
& \texttt{Qwen2.5 7B}   & 0.679 & 0.652 & 0.546 & 0.416 \\
\midrule
\multirow{4}{*}{Phone}
& \texttt{Llama 3.2 1B} & 0.380 & 0.293 & 0.389 & 0.254 \\
& \texttt{Llama 3.2 3B} & 0.398 & 0.279 & 0.372 & 0.243 \\
& \texttt{Qwen2.5 3B}   & 0.714 & 0.365 & 0.652 & 0.434 \\
& \texttt{Qwen2.5 7B}   & 0.695 & 0.369 & 0.619 & 0.399 \\
\bottomrule
\end{tabular}
}
\end{table}

\paragraph{The predicted ordering holds for Twitter handles and emails, and for phone numbers except in German}
As we discussed in Section~\ref{sec:finding-bridges}, if the models bridge
different languages' representations, three conditions should hold: strict
translation should preserve greater similarity to the English original than
an English paraphrase (H1: Equation~\ref{eq:L1-paraL1}), an English
paraphrase should preserve greater similarity than a target-language
paraphrase (H2: Equation~\ref{eq:paraL1-paraL2}), and strict translation
should consequently exceed the target-language paraphrase as well
(H3: Equation~\ref{eq:L1-paraL2}). For Twitter handles and emails, all three hold for every one of the four models, for both the French/Italian/Spanish translations and the German one, when similarity is averaged over layers: strict translation $>$ English paraphrase $>$ target language paraphrase in every row of Table~\ref{tab:bridge_summary}. For phone numbers, H2 holds in every row, while H1 and H3 hold for French, Italian and Spanish (for H1, in three of the four models; Llama 3.2 1B is on par, 0.380 vs 0.389). German is the exception: its strict translation is below the English paraphrase at every layer and in every model, and, for Qwen2.5, below the average target-language paraphrase as well (0.365 vs 0.434 and 0.369 vs 0.399). Where it holds, this ordering is consistent with \textbf{a bridge anchored to content fidelity}: representational similarity degrades least when only the
language changes (strict translation), more when only the wording changes
(English paraphrase), and most when both change simultaneously
(target language paraphrase).

\paragraph{Multilingual models show a substantially stronger bridge than English-centric ones}
The gap between conditions and the absolute similarity level scale
with the model's multilingual training. The two Llama models -- despite being described as multilingual -- stay below 0.4 mean cosine similarity in every condition, and the layer-averaged difference between strict translation (French/Italian/Spanish) and target-language paraphrase is comparatively narrow (0.09--0.16). Qwen2.5, trained on a substantially more multilingual corpus, reaches mean similarities of 0.67--0.71 for the strict condition and shows a much wider separation between conditions (a Strict/target language paraphrase gap of roughly 0.23--0.30). This mirrors the extraction-transferability results in
Section~\ref{sec:transferability-results}: the same models whose leak rates
transfer most completely across languages are also the ones whose internal
representations show the clearest content-anchored bridge.

\paragraph{The bridge is present for all three PII types, but its strength does not follow the leakage ordering}
The Strict/target language paraphrase gap (French/Italian/Spanish strict translations) is 0.13--0.29 for Twitter handles, 0.09--0.26 for emails and 0.13--0.30 for phone numbers (Table~\ref{tab:bridge_summary}). The gap is smallest for emails in every model, and comparable between Twitter handles and phone numbers. Unlike the leakage rates presented in Section~\ref{sec:transferability-results}, 
where Twitter handles are leaked most and phone numbers least, the strength of the bridge therefore does not order the PII types consistently, and we do not read it as a proxy for extraction risk across PII types.

Twitter handles also show a distinct pattern: target-language paraphrases sit closer to each other (0.04--0.11 apart) than either does to strict translation (0.09--0.29 apart).

German phone numbers are the exception to the ordering above: their strict translation reaches only 0.28--0.37 mean similarity (versus 0.38--0.71 for French, Italian and Spanish), it lies below the English paraphrase at every layer, and it is the only strict condition that falls below the target-language paraphrase in some models. This coincides with the sharp drop in the number of generations containing a PII and of leaks for German phone numbers (Table~\ref{tab:leakage_counts}), suggesting that this condition behaves differently from the others; we therefore report it separately in Table~\ref{tab:bridge_summary}.

\paragraph{Representations are bridged in the middle layers, and then diverge}
Across all four models, the layer-wise curves trace a consistent arc:
similarity to the English original is lowest at the input end (and, for Qwen2.5, also at the output end), and highest in between, where the cross-lingual bridge appears
to be located. At the embedding layer (layer~0), before any contextual
processing has occurred, similarity is driven mostly by shared surface
tokens. For Twitter handles and emails, the predicted ordering (strict $>$ eng-paraphrase $>$ L2-paraphrase) reasserts itself within the first few transformer layers and holds through the middle of the network (in at least 97\% of the middle layers), which we take as the region where content-level, language-agnostic representations are formed. For phone numbers, this ordering holds less consistently (75--96\% of the middle layers for French, Italian, and Spanish), while the German strict translation stays below the English paraphrase at every layer.

Toward the output end, the bridge again gives way: for Qwen2.5 in particular, all three conditions -- strict, eng-paraphrase, and L2-paraphrase alike -- show a pronounced drop in absolute similarity over the final one to three layers (a fall of roughly 0.12--0.35 in cosine similarity from each curve's peak). We read this late-layer divergence as a property of the output layers generally, not of the bridge specifically: the last layers of a language model are known to specialize toward predicting the next token in the correct surface language, which pulls every condition's representation away from the English original by a similar amount. Critically, the target-language paraphrase remains the furthest from the English original at every layer, all the way to the final one, while the strict translation and the English paraphrase come close to each other (at the very last layer the English paraphrase is on par with or slightly above the strict translation for emails and phone numbers). The late-layer drop therefore rescales the overall level of similarity without erasing the separation between the target-language paraphrase and the other conditions, which is part of the evidence for the bridge. Taken together, the arc suggests that cross-lingual leakage is enabled by a mid-network representation that is comparatively insulated from both the surface form of the input and the language-specific demands of output generation.

The representational analysis corroborates the attack-level transferability results: a language-crossing bridge exists, it is anchored to content fidelity instead of topic alone (for Twitter handles, emails and French/Italian/Spanish phone numbers, translation beats paraphrase, even a same-language one), and its strength scales with a model's multilingual training, not its parameter count. Its strength across PII types, however, does not mirror the leakage counts in Table~\ref{tab:leakage_counts}, and the German phone-number condition remains an open exception.

\section{Discussion and Conclusions}
This paper set out to test an assumption implicit in most prior work on
training data extraction: that an attacker queries a model in the same
language the memorized content was originally written in. Our results
show this assumption does not hold. TDE attacks transfer across the
language boundary for both English-centric and multilingual models, a
sample of the translated passages (200 per PII type) was verified to be
absent from the web prior to our experiments, and the degree of transfer
tracks a model's multilingual training.

By independently varying language and wording fidelity, we further showed that
this leakage is anchored to the specific memorized content: a strict translation recovers substantially more leakage than a paraphrase in either English or the target language, and, for Twitter handles and emails, also produces a more similar internal representation (for phone numbers, this holds for French, Italian and Spanish but not for German).

Representationally, this content-anchored bridge is neither present from the input embeddings nor preserved into the output layers, but is concentrated in a band of middle transformer layers, consistent with the view that these layers encode language-agnostic, fact-level associations before the final layers specialize toward generation in the surface language of the prompt \cite{wendler-etal-2024-llamas}.

\paragraph{Connection to cross-lingual unlearning} These results offer a
plausible mechanistic account for a finding we could previously only cite,
not explain: \cite{choi-etal-2024-cross} show that unlearning a fact in one
language does not remove it in another -- a deletion in English typically
leaves the same fact recoverable, verbatim, in Spanish or Chinese. If the
cross-lingual bridge we identify is concentrated in the middle layers, while
existing unlearning and model-editing methods largely target specific factual associations
localized around particular parameters or layers, it is plausible that such
methods remove the language-specific surface association without touching
the shared mid-layer representation that the bridge depends on. Under this
account, cross-lingual leakage after unlearning would not be a failure of
unlearning \emph{per se}, but a symptom of editing a representation at the
wrong depth. We see directly testing this account -- by applying unlearning
or editing methods and then repeating our layer-wise similarity analysis --
as a natural and high-value extension of this work.

\paragraph{Practical implications} For practitioners, the central
implication is that monolingual privacy evaluation is not sufficient for a
multilingual model. A model that appears to resist an English-language
extraction attempt cannot be assumed to resist the same attempt translated
into another language, and our results suggest the opposite is more likely
as multilingual training becomes more extensive. This has direct bearing on
red-teaming and compliance practice: privacy audits, safety filters, and
memorization benchmarks that are evaluated only in a model's dominant
language will systematically underestimate real-world exposure for any
model with meaningful multilingual capability -- which, at this point,
describes most production-grade LLMs.

\paragraph{Limitations} Our target languages -- French, Italian, Spanish,
and German -- are all high-resource and, with the partial exception of
German, typologically close to English. This was a deliberate choice to
isolate the language-versus-wording effect under favorable translation
quality, but it leaves open, and likely understates, the risk for
lower-resource or typologically distant languages, where prior work
suggests both weaker safety alignment \cite{Li2024CrossLanguage} and
different memorization dynamics \cite{luo-etal-2025-shared,
mishra-etal-2026-models, satvaty-etal-2025-memorization} may amplify instead of
dampen the effect we observe. Second, our phone-number results rest on small absolute leak
counts, particularly under paraphrase and for the English-centric models,
and should be interpreted with corresponding caution. In particular, the German phone-number condition behaves differently from the other languages, both behaviorally (Table~\ref{tab:leakage_counts}) and representationally (Table~\ref{tab:bridge_summary}, Fig.~\ref{fig:bridge}), and we report it separately.

\paragraph{Future work.} Beyond directly testing the unlearning connection
above, three extensions follow naturally from our limitations. First,
repeating this study on lower-resource and typologically distant target
languages would test whether the cross-lingual bridge strengthens,
weakens, or changes character outside the high-resource, Western-European
setting studied here. Second, extending the paraphrase-based
disentanglement methodology from extraction to jailbreaking would let us
test whether the cross-lingual latent space that \cite{Li2024CrossLanguage}
exploit for safety circumvention has the same content-anchored,
middle-layer structure we find for memorization, or whether the two
phenomena rely on different mechanisms despite superficially similar
symptoms. Third, our similarity-based evidence for the bridge is
correlational at the representation level; combining it with causal
interventions -- for instance, patching or ablating the token-level
contributions we identify as most responsible for cross-lingual similarity
-- would let us move from showing that a bridge exists to showing precisely
which components of the model implement it.

In summary, we provide the first direct evidence that PII memorized from
English training data can be recovered by attacks formulated in other
languages, that this transfer is driven by a genuine, content-anchored
cross-lingual bridge and that the
strength of this bridge scales with a model's multilingual training. As
multilingual deployment continues to expand, we argue that privacy
evaluation, auditing, and mitigation must be treated as inherently
cross-lingual problems, not extended to other languages as an afterthought.

\section*{Open Science}
We release the full code required to reproduce our experiments (the TDE
attack pipeline, the translation and paraphrase generation scripts, the
web-presence verification pipeline, and the analysis and plotting notebooks
used to produce every table and figure in this paper) via a fully
anonymized repository: \url{https://anonymous.4open.science/r/TransferTDE-CC84/}. This repository will remain accessible throughout the review process.
We do not release the underlying dataset itself -- the English passages, and
their translations and paraphrases -- as it consists of samples containing
real, identifiable personal information (email addresses, phone numbers,
and Twitter handles) drawn from \texttt{Pile-CC}, and redistributing it
would work against the privacy considerations (discussed in our Ethical
Considerations section). We will make the constructed dataset available to researchers upon request for legitimate research purposes, subject to a data use agreement and institutional verification.

\section*{LLM usage considerations}
\setcounter{paragraph}{0}
\paragraph{Accountability and Correctness}
LLMs were used for editorial purposes in this manuscript, and all outputs were inspected by the authors to ensure accuracy and originality. 
The authors reviewed and edited all LLM-generated content used in the paper,
including generated code, as if it had been written by the authors themselves,
and take full responsibility for its accuracy and integrity.
Beyond editorial use, an LLM (Gemma-3-27B-it, accessed via the Google
Gemini API) was used methodologically, as a core component of our data
pipeline, to generate the translations described in
Section~\ref{sec:data}. The paraphrases used in
Section~\ref{sec:finding-bridges} were instead generated with open-weight models
run locally (Appendix~\ref{app:paraphrase}).

At least one author proficient in each target language conducted spot checks of the translations and paraphrases, and we additionally
verified the integrity of the paraphrases in Appendix ~\ref{app:paraphrase}.
No experimental results, references, or claims in this paper were generated by an LLM without independent verification against the underlying data or source material.

\paragraph{Transparency}
The use of an LLM to generate translations is an integral part of our
methodology, not a secondary element: our experimental design relies on
the availability of faithful, structure-preserving translations of a
dataset of the required scale, which we obtained by instructing
Gemma-3-27B-it to that effect (Section~\ref{sec:data}). The
free-form paraphrases, which provide the contrasting fidelity condition
(Section~\ref{sec:finding-bridges}), were generated with open-weight
models run locally (Appendix~\ref{app:paraphrase}). We independently
validated the results of this process. We acknowledge a limitation that
this choice introduces: since Gemma-3-27B-it is accessed via a hosted
API with versioned weights, our translation generation phase is not
perfectly reproducible if the underlying model provided by the API
undergoes changes or is deprecated in the future. The paraphrases are
not affected by this limitation, as they were produced with locally-run
open-weight models. We mitigate this issue by publishing, in our
artifacts, the exact prompts and scripts used to generate the
translations and paraphrases in this article (subject to the
data-sharing constraints described in our Open Science section).

\paragraph{Responsibility}
The target models evaluated in this study (GPT-Neo~1.3B/2.7B, GPT-J~6B, Llama~3.2~1B/3B, Qwen2.5~3B/7B) are the \emph{subjects} of our study; we did not train any models, nor did we collect new data beyond the translation and paraphrasing of pre-existing texts from \texttt{Pile-CC} (Section~\ref{sec:data}).
We selected models covering two parameter ranges (approximately 1 to 3B and 6 to 7B) within each family specifically to verify whether cross-lingual transfer is proportional to model size or to the composition of the multilingual pre-training; our results point to the latter hypothesis, which justified our decision to limit our evaluation to this range and not to more computationally intensive models.
All inference on the target models was performed using greedy decoding, limiting each query to a single forward generation step. The dataset used by PII type and language was restricted to obtain statistically significant leakage counts without querying the models more than necessary for this purpose. The experiments were run on a single RTX A6000 from a private, local server. Each inference took from 10s/it for smaller models up to 20s/it for larger ones. Considering also the large number of hours needed to perfect the experimental setup, we estimate a total of 100 hours of GPU usage. The estimated emissions are 12.96\,kg of CO\textsubscript{2}.\footnote{Estimate computed on \url{https://mlco2.github.io/impact/}}

\section*{Ethical Considerations}
\setcounter{paragraph}{0}
This paper analyzes whether personally identifiable information (PII)
stored by large language models during pre-training can be extracted
using prompts formulated in a language other than the one in which the
information was originally recorded. Below, we discuss the risks this entails
and the measures we have taken to mitigate them.

\paragraph{Data provenance and handling} All PII used in our experiments comes from \texttt{Pile-CC}, a pre-existing dataset derived from Common Crawl. We did not collect, extract, or generate any new PII; the email addresses, phone numbers, and Twitter handles we analyze were already present in this corpus and, by extension, already present in the training data of the models we evaluate, prior to our intervention. Our translations and paraphrases are transformations of this existing data, not new instances of it. We did not attempt to identify, contact, or otherwise interact with the individuals associated with the PII in our dataset, and we report our results in aggregate form (number of data leaks and similarity statistics). 
The English extracts, which contain PII, were processed by two
third-party services: (i) the Google Gemini API, to which we sent the
extracts to obtain the translations (Section~\ref{sec:data}), and
(ii) the Tavily search API, to which we sent queries derived from the
translated extracts for the web-presence check
(Section~\ref{sec:novelty}). No PII was sent to third-party services
for paraphrase generation, which was performed locally
(Appendix~\ref{app:paraphrase}).
We do not publish, list, or otherwise reproduce any of the individual instances of PII retrieved during our experiments. Such information might be shared for research purposes only by contacting directly the corresponding authors.

\paragraph{Dual-use considerations}
Attacks based on the extraction of training data are, by their very nature, dual-use: the same techniques that allow us to measure a privacy risk could, in principle, be used by a malicious actor to materialize that risk. We believe that disclosing the results we have obtained promotes the development of new defense techniques for two reasons. 
First, the underlying vulnerability (monolingual memorization and verbatim retrieval) has already been well established in previous work; our contribution lies in demonstrating that current monolingual extraction and evaluation practices are insufficient, which in itself constitutes an argument in favor of more thorough defense and oversight.
Second, concealing this finding would prevent model providers and downstream implementers from realizing that monolingual privacy evaluation, red-teaming, and mitigation measures (e.g., unlearning or differential privacy applied and tested in only one language) may substantially underestimate real-world exposure, which, in our view, represents a greater harm than the marginal advantage our results provide to an attacker who could, in any case, already translate a known extraction prompt on their own.

\paragraph{Responsible disclosure}
We did not pursue formal disclosure, as all the models evaluated are open-weight, and our findings pertain to a general property of multilingual training. We encourage practitioners who apply unlearning techniques, differential privacy, or other mitigation measures to evaluate their effectiveness across languages.

\paragraph{Scope of risk}
The types of PII we study (email addresses, phone numbers, and Twitter handles) are all categories of information that may already be publicly available on the web in the specific cases we extracted from Pile-CC, and our web presence check (Section~\ref{sec:novelty}) was designed to verify the novelty of the translation for the purposes of our methodology, not to assess whether the underlying English-language PII remains publicly identifiable through other means. We did not evaluate the extraction of more sensitive categories of PII (e.g., government-issued identifiers, financial or health information), and we caution against extrapolating our leakage rate estimates to such categories without a separate evaluation.

\bibliographystyle{IEEEtran}
\bibliography{custom}

@inproceedings{papineni2002bleu,
   title={BLEU: a method for automatic evaluation of machine translation},
   author={Papineni, Kishore and Roukos, Salim and Ward, Todd and Zhu, Wei-Jing},
   booktitle={Proceedings of the 40th Annual Meeting of the ACL},
   year={2002}
}

@inproceedings{lin2004rouge, title={ROUGE: A Package for Automatic Evaluation of Summaries},
   author={Lin, Chin-Yew},
   booktitle={Text Summarization Branches Out},
   year={2004}
}

@inproceedings{conneau2020xnli,
   title={Unsupervised Cross-lingual Representation Learning at Scale},
   author={Conneau, Alexis and others},
   booktitle={Proceedings of ACL},
   year={2020}
}

@misc{nasr2023scalableextractiontrainingdata,
      title={Scalable Extraction of Training Data from (Production) Language Models}, 
      author={Milad Nasr and Nicholas Carlini and Jonathan Hayase and Matthew Jagielski and A. Feder Cooper and Daphne Ippolito and Christopher A. Choquette-Choo and Eric Wallace and Florian Tramèr and Katherine Lee},
      year={2023},
      eprint={2311.17035},
      archivePrefix={arXiv},
      primaryClass={cs.LG},
      url={https://arxiv.org/abs/2311.17035}, 
}

@inproceedings{huang-etal-2022-large,
    title = "Are Large Pre-Trained Language Models Leaking Your Personal Information?",
    author = "Huang, Jie  and
      Shao, Hanyin  and
      Chang, Kevin Chen-Chuan",
    editor = "Goldberg, Yoav  and
      Kozareva, Zornitsa  and
      Zhang, Yue",
    booktitle = "Findings of the Association for Computational Linguistics: EMNLP 2022",
    month = dec,
    year = "2022",
    address = "Abu Dhabi, United Arab Emirates",
    publisher = "Association for Computational Linguistics",
    url = "https://aclanthology.org/2022.findings-emnlp.148/",
    doi = "10.18653/v1/2022.findings-emnlp.148",
    pages = "2038--2047"
}

@misc{gao2020pile800gbdatasetdiverse,
      title={The Pile: An 800GB Dataset of Diverse Text for Language Modeling}, 
      author={Leo Gao and Stella Biderman and Sid Black and Laurence Golding and Travis Hoppe and Charles Foster and Jason Phang and Horace He and Anish Thite and Noa Nabeshima and Shawn Presser and Connor Leahy},
      year={2020},
      eprint={2101.00027},
      archivePrefix={arXiv},
      primaryClass={cs.CL},
      url={https://arxiv.org/abs/2101.00027}, 
}

@misc{srivastava2025owlprobingcrosslingualrecall,
      title={OWL: Probing Cross-Lingual Recall of Memorized Texts via World Literature}, 
      author={Alisha Srivastava and Emir Korukluoglu and Minh Nhat Le and Duyen Tran and Chau Minh Pham and Marzena Karpinska and Mohit Iyyer},
      year={2025},
      eprint={2505.22945},
      archivePrefix={arXiv},
      primaryClass={cs.CL},
      url={https://arxiv.org/abs/2505.22945}, 
}

@inproceedings{choi-etal-2024-cross,
    title = "Cross-Lingual Unlearning of Selective Knowledge in Multilingual Language Models",
    author = "Choi, Minseok  and
      Min, Kyunghyun  and
      Choo, Jaegul",
    editor = "Al-Onaizan, Yaser  and
      Bansal, Mohit  and
      Chen, Yun-Nung",
    booktitle = "Findings of the Association for Computational Linguistics: EMNLP 2024",
    month = nov,
    year = "2024",
    address = "Miami, Florida, USA",
    publisher = "Association for Computational Linguistics",
    url = "https://aclanthology.org/2024.findings-emnlp.630/",
    doi = "10.18653/v1/2024.findings-emnlp.630",
    pages = "10732--10747"
}

@inproceedings{mishra-etal-2026-models,
    title = "The Model{'}s Language Matters: A Comparative Privacy Analysis of {LLM}s",
    author = "Mishra, Abhishek Kumar  and
      Boutet, Antoine  and
      Magnana, Lucas",
    editor = "Demberg, Vera  and
      Inui, Kentaro  and
      Marquez, Llu{\'i}s",
    booktitle = "Findings of the {A}ssociation for {C}omputational {L}inguistics: {EACL} 2026",
    month = mar,
    year = "2026",
    address = "Rabat, Morocco",
    publisher = "Association for Computational Linguistics",
    url = "https://aclanthology.org/2026.findings-eacl.54/",
    doi = "10.18653/v1/2026.findings-eacl.54",
    pages = "1038--1048",
    ISBN = "979-8-89176-386-9"
}

@inproceedings{satvaty-etal-2025-memorization,
    title = "Memorization is Language-Sensitive: Analyzing Memorization and Inference Risks of {LLM}s in a Multilingual Setting",
    author = "Satvaty, Ali  and
      Visman, Anna  and
      Seidel, Daniel  and
      Verberne, Suzan  and
      Turkmen, Fatih",
    editor = "Jia, Robin  and
      Wallace, Eric  and
      Huang, Yangsibo  and
      Pimentel, Tiago  and
      Maini, Pratyush  and
      Dankers, Verna  and
      Wei, Johnny  and
      Lesci, Pietro",
    booktitle = "Proceedings of the First Workshop on Large Language Model Memorization (L2M2)",
    month = aug,
    year = "2025",
    address = "Vienna, Austria",
    publisher = "Association for Computational Linguistics",
    url = "https://aclanthology.org/2025.l2m2-1.9/",
    doi = "10.18653/v1/2025.l2m2-1.9",
    pages = "106--126",
    ISBN = "979-8-89176-278-7"
}

@inproceedings{luo-etal-2025-shared,
    title = "Shared Path: Unraveling Memorization in Multilingual {LLM}s through Language Similarities",
    author = "Luo, Xiaoyu  and
      Chen, Yiyi  and
      Bjerva, Johannes  and
      Li, Qiongxiu",
    editor = "Christodoulopoulos, Christos  and
      Chakraborty, Tanmoy  and
      Rose, Carolyn  and
      Peng, Violet",
    booktitle = "Proceedings of the 2025 Conference on Empirical Methods in Natural Language Processing",
    month = nov,
    year = "2025",
    address = "Suzhou, China",
    publisher = "Association for Computational Linguistics",
    url = "https://aclanthology.org/2025.emnlp-main.978/",
    doi = "10.18653/v1/2025.emnlp-main.978",
    pages = "19372--19388",
    ISBN = "979-8-89176-332-6"
}

@inproceedings{ICLR2026_5f124a19,
 author = {Nikolaou, Giorgos and Mencattini, Tommaso and Crisostomi, Donato and Santilli, Andrea and Panagakis, Yannis and Rodol\`{a}, Emanuele},
 booktitle = {International Conference on Learning Representations},
 editor = {C. Vondrick and B. Hariharan and C. Raffel and L. Pinto and D. Yang and A. Faust},
 pages = {57927--57984},
 title = {Language Models are Injective and Hence Invertible},
 url = {https://proceedings.iclr.cc/paper_files/paper/2026/file/5f124a1979570872ff46a6de8fbac35c-Paper-Conference.pdf},
 volume = {2026},
 year = {2026}
}

@inproceedings{wendler-etal-2024-llamas,
    title = "Do Llamas Work in {E}nglish? On the Latent Language of Multilingual Transformers",
    author = "Wendler, Chris  and
      Veselovsky, Veniamin  and
      Monea, Giovanni  and
      West, Robert",
    editor = "Ku, Lun-Wei  and
      Martins, Andre  and
      Srikumar, Vivek",
    booktitle = "Proceedings of the 62nd Annual Meeting of the Association for Computational Linguistics (Volume 1: Long Papers)",
    month = aug,
    year = "2024",
    address = "Bangkok, Thailand",
    publisher = "Association for Computational Linguistics",
    url = "https://aclanthology.org/2024.acl-long.820/",
    doi = "10.18653/v1/2024.acl-long.820",
    pages = "15366--15394"
}

@Article{venditti2026pae,
AUTHOR = {Venditti, Davide and Ruzzetti, Elena Sofia and Xompero, Giancarlo A. and Giannone, Cristina and Favalli, Andrea and Romagnoli, Raniero and Zanzotto, Fabio Massimo},
TITLE = {Enhancing Data Privacy in Large Language Models Through Private Association Editing},
JOURNAL = {Journal of Cybersecurity and Privacy},
VOLUME = {6},
YEAR = {2026},
NUMBER = {3},
ARTICLE-NUMBER = {96},
URL = {https://www.mdpi.com/2624-800X/6/3/96},
ISSN = {2624-800X},
DOI = {10.3390/jcp6030096}
}

@inproceedings{carlini2021extracting,
  author    = {Nicholas Carlini and Florian Tram{\`e}r and Eric Wallace and
               Matthew Jagielski and Ariel Herbert-Voss and Katherine Lee and
               Adam Roberts and Tom Brown and Dawn Song and
               {\'U}lfar Erlingsson and Alina Oprea and Colin Raffel},
  title     = {Extracting Training Data from Large Language Models},
  booktitle = {30th USENIX Security Symposium (USENIX Security 21)},
  pages     = {2633--2650},
  year      = {2021},
  publisher = {USENIX Association},
  url       = {https://www.usenix.org/conference/usenixsecurity21/presentation/carlini-extracting}
}

@article{Li2024CrossLanguage,
  author={Li, Jie and Shi, Ling and Liu, Yi and Ren, Xiaoning and Liu, Yang},
  title={A Cross-Language Investigation into Jailbreak Attacks in Large Language Models},
  journal={arXiv preprint arXiv:2401.16765},
  year={2024},
}

@inproceedings{ruzzetti2025pme,
    title = "Private Memorization Editing: Turning Memorization into a Defense to Strengthen Data Privacy in Large Language Models",
    author = "Ruzzetti, Elena Sofia  and
      Xompero, Giancarlo A.  and
      Venditti, Davide  and
      Zanzotto, Fabio Massimo",
    editor = "Che, Wanxiang  and
      Nabende, Joyce  and
      Shutova, Ekaterina  and
      Pilehvar, Mohammad Taher",
    booktitle = "Proceedings of the 63rd Annual Meeting of the Association for Computational Linguistics (Volume 1: Long Papers)",
    month = jul,
    year = "2025",
    address = "Vienna, Austria",
    publisher = "Association for Computational Linguistics",
    url = "https://aclanthology.org/2025.acl-long.810/",
    doi = "10.18653/v1/2025.acl-long.810",
    pages = "16572--16592",
    ISBN = "979-8-89176-251-0"
}

@misc{carlini2024stealingproductionlanguagemodel,
      title={Stealing Part of a Production Language Model}, 
      author={Nicholas Carlini and Daniel Paleka and Krishnamurthy Dj Dvijotham and Thomas Steinke and Jonathan Hayase and A. Feder Cooper and Katherine Lee and Matthew Jagielski and Milad Nasr and Arthur Conmy and Itay Yona and Eric Wallace and David Rolnick and Florian Tramèr},
      year={2024},
      eprint={2403.06634},
      archivePrefix={arXiv},
      primaryClass={cs.CR},
      url={https://arxiv.org/abs/2403.06634}, 
}

@Article{Hoory2021DP,
AUTHOR = {Kantheti, Bhagiradh and Araujo, Carlos A. Paz De},
TITLE = {Privacy-Preserving Transformers for Time-Series Forecasting: A Survey and Two-Axis Taxonomy of Cryptographic, Differential-Privacy, and Feature-Path Methods},
JOURNAL = {AI},
VOLUME = {7},
YEAR = {2026},
NUMBER = {9},
ARTICLE-NUMBER = {340},
URL = {https://www.mdpi.com/2673-2688/7/9/340},
ISSN = {2673-2688},
DOI = {10.3390/ai7090340}
}

@misc{gpt-j,
  author = {Wang, Ben and Komatsuzaki, Aran},
  title = {{GPT-J-6B: A 6 Billion Parameter Autoregressive Language Model}},
  howpublished = {\url{https://github.com/kingoflolz/mesh-transformer-jax}},
  year = 2021,
  month = May
}

@misc{qwen2025qwen25technicalreport,
      title={Qwen2.5 Technical Report}, 
      author={Qwen and : and An Yang and Baosong Yang and Beichen Zhang and Binyuan Hui and Bo Zheng and Bowen Yu and Chengyuan Li and Dayiheng Liu and Fei Huang and Haoran Wei and Huan Lin and Jian Yang and Jianhong Tu and Jianwei Zhang and Jianxin Yang and Jiaxi Yang and Jingren Zhou and Junyang Lin and Kai Dang and Keming Lu and Keqin Bao and Kexin Yang and Le Yu and Mei Li and Mingfeng Xue and Pei Zhang and Qin Zhu and Rui Men and Runji Lin and Tianhao Li and Tianyi Tang and Tingyu Xia and Xingzhang Ren and Xuancheng Ren and Yang Fan and Yang Su and Yichang Zhang and Yu Wan and Yuqiong Liu and Zeyu Cui and Zhenru Zhang and Zihan Qiu},
      year={2025},
      eprint={2412.15115},
      archivePrefix={arXiv},
      primaryClass={cs.CL},
      url={https://arxiv.org/abs/2412.15115}, 
}

@misc{grattafiori2024llama3herdmodels,
      title={The Llama 3 Herd of Models}, 
      author={Aaron Grattafiori and Abhimanyu Dubey and Abhinav Jauhri and Abhinav Pandey and Abhishek Kadian and Ahmad Al-Dahle and Aiesha Letman and Akhil Mathur and Alan Schelten and Alex Vaughan and Amy Yang and Angela Fan and Anirudh Goyal and Anthony Hartshorn and Aobo Yang and Archi Mitra and Archie Sravankumar and Artem Korenev and Arthur Hinsvark and Arun Rao and Aston Zhang and Aurelien Rodriguez and Austen Gregerson and Ava Spataru and Baptiste Roziere and Bethany Biron and Binh Tang and Bobbie Chern and Charlotte Caucheteux and Chaya Nayak and Chloe Bi and Chris Marra and Chris McConnell and Christian Keller and Christophe Touret and Chunyang Wu and Corinne Wong and Cristian Canton Ferrer and Cyrus Nikolaidis and Damien Allonsius and Daniel Song and Danielle Pintz and Danny Livshits and Danny Wyatt and David Esiobu and Dhruv Choudhary and Dhruv Mahajan and Diego Garcia-Olano and Diego Perino and Dieuwke Hupkes and Egor Lakomkin and Ehab AlBadawy and Elina Lobanova and Emily Dinan and Eric Michael Smith and Filip Radenovic and Francisco Guzmán and Frank Zhang and Gabriel Synnaeve and Gabrielle Lee and Georgia Lewis Anderson and Govind Thattai and Graeme Nail and Gregoire Mialon and Guan Pang and Guillem Cucurell and Hailey Nguyen and Hannah Korevaar and Hu Xu and Hugo Touvron and Iliyan Zarov and Imanol Arrieta Ibarra and Isabel Kloumann and Ishan Misra and Ivan Evtimov and Jack Zhang and Jade Copet and Jaewon Lee and Jan Geffert and Jana Vranes and Jason Park and Jay Mahadeokar and Jeet Shah and Jelmer van der Linde and Jennifer Billock and Jenny Hong and Jenya Lee and Jeremy Fu and Jianfeng Chi and Jianyu Huang and Jiawen Liu and Jie Wang and Jiecao Yu and Joanna Bitton and Joe Spisak and Jongsoo Park and Joseph Rocca and Joshua Johnstun and Joshua Saxe and Junteng Jia and Kalyan Vasuden Alwala and Karthik Prasad and Kartikeya Upasani and Kate Plawiak and Ke Li and Kenneth Heafield and Kevin Stone and Khalid El-Arini and Krithika Iyer and Kshitiz Malik and Kuenley Chiu and Kunal Bhalla and Kushal Lakhotia and Lauren Rantala-Yeary and Laurens van der Maaten and Lawrence Chen and Liang Tan and Liz Jenkins and Louis Martin and Lovish Madaan and Lubo Malo and Lukas Blecher and Lukas Landzaat and Luke de Oliveira and Madeline Muzzi and Mahesh Pasupuleti and Mannat Singh and Manohar Paluri and Marcin Kardas and Maria Tsimpoukelli and Mathew Oldham and Mathieu Rita and Maya Pavlova and Melanie Kambadur and Mike Lewis and Min Si and Mitesh Kumar Singh and Mona Hassan and Naman Goyal and Narjes Torabi and Nikolay Bashlykov and Nikolay Bogoychev and Niladri Chatterji and Ning Zhang and Olivier Duchenne and Onur Çelebi and Patrick Alrassy and Pengchuan Zhang and Pengwei Li and Petar Vasic and Peter Weng and Prajjwal Bhargava and Pratik Dubal and Praveen Krishnan and Punit Singh Koura and Puxin Xu and Qing He and Qingxiao Dong and Ragavan Srinivasan and Raj Ganapathy and Ramon Calderer and Ricardo Silveira Cabral and Robert Stojnic and Roberta Raileanu and Rohan Maheswari and Rohit Girdhar and Rohit Patel and Romain Sauvestre and Ronnie Polidoro and Roshan Sumbaly and Ross Taylor and Ruan Silva and Rui Hou and Rui Wang and Saghar Hosseini and Sahana Chennabasappa and Sanjay Singh and Sean Bell and Seohyun Sonia Kim and Sergey Edunov and Shaoliang Nie and Sharan Narang and Sharath Raparthy and Sheng Shen and Shengye Wan and Shruti Bhosale and Shun Zhang and Simon Vandenhende and Soumya Batra and Spencer Whitman and Sten Sootla and Stephane Collot and Suchin Gururangan and Sydney Borodinsky and Tamar Herman and Tara Fowler and Tarek Sheasha and Thomas Georgiou and Thomas Scialom and Tobias Speckbacher and Todor Mihaylov and Tong Xiao and Ujjwal Karn and Vedanuj Goswami and Vibhor Gupta and Vignesh Ramanathan and Viktor Kerkez and Vincent Gonguet and Virginie Do and Vish Vogeti and Vítor Albiero and Vladan Petrovic and Weiwei Chu and Wenhan Xiong and Wenyin Fu and Whitney Meers and Xavier Martinet and Xiaodong Wang and Xiaofang Wang and Xiaoqing Ellen Tan and Xide Xia and Xinfeng Xie and Xuchao Jia and Xuewei Wang and Yaelle Goldschlag and Yashesh Gaur and Yasmine Babaei and Yi Wen and Yiwen Song and Yuchen Zhang and Yue Li and Yuning Mao and Zacharie Delpierre Coudert and Zheng Yan and Zhengxing Chen and Zoe Papakipos and Aaditya Singh and Aayushi Srivastava and Abha Jain and Adam Kelsey and Adam Shajnfeld and Adithya Gangidi and Adolfo Victoria and Ahuva Goldstand and Ajay Menon and Ajay Sharma and Alex Boesenberg and Alexei Baevski and Allie Feinstein and Amanda Kallet and Amit Sangani and Amos Teo and Anam Yunus and Andrei Lupu and Andres Alvarado and Andrew Caples and Andrew Gu and Andrew Ho and Andrew Poulton and Andrew Ryan and Ankit Ramchandani and Annie Dong and Annie Franco and Anuj Goyal and Aparajita Saraf and Arkabandhu Chowdhury and Ashley Gabriel and Ashwin Bharambe and Assaf Eisenman and Azadeh Yazdan and Beau James and Ben Maurer and Benjamin Leonhardi and Bernie Huang and Beth Loyd and Beto De Paola and Bhargavi Paranjape and Bing Liu and Bo Wu and Boyu Ni and Braden Hancock and Bram Wasti and Brandon Spence and Brani Stojkovic and Brian Gamido and Britt Montalvo and Carl Parker and Carly Burton and Catalina Mejia and Ce Liu and Changhan Wang and Changkyu Kim and Chao Zhou and Chester Hu and Ching-Hsiang Chu and Chris Cai and Chris Tindal and Christoph Feichtenhofer and Cynthia Gao and Damon Civin and Dana Beaty and Daniel Kreymer and Daniel Li and David Adkins and David Xu and Davide Testuggine and Delia David and Devi Parikh and Diana Liskovich and Didem Foss and Dingkang Wang and Duc Le and Dustin Holland and Edward Dowling and Eissa Jamil and Elaine Montgomery and Eleonora Presani and Emily Hahn and Emily Wood and Eric-Tuan Le and Erik Brinkman and Esteban Arcaute and Evan Dunbar and Evan Smothers and Fei Sun and Felix Kreuk and Feng Tian and Filippos Kokkinos and Firat Ozgenel and Francesco Caggioni and Frank Kanayet and Frank Seide and Gabriela Medina Florez and Gabriella Schwarz and Gada Badeer and Georgia Swee and Gil Halpern and Grant Herman and Grigory Sizov and Guangyi and Zhang and Guna Lakshminarayanan and Hakan Inan and Hamid Shojanazeri and Han Zou and Hannah Wang and Hanwen Zha and Haroun Habeeb and Harrison Rudolph and Helen Suk and Henry Aspegren and Hunter Goldman and Hongyuan Zhan and Ibrahim Damlaj and Igor Molybog and Igor Tufanov and Ilias Leontiadis and Irina-Elena Veliche and Itai Gat and Jake Weissman and James Geboski and James Kohli and Janice Lam and Japhet Asher and Jean-Baptiste Gaya and Jeff Marcus and Jeff Tang and Jennifer Chan and Jenny Zhen and Jeremy Reizenstein and Jeremy Teboul and Jessica Zhong and Jian Jin and Jingyi Yang and Joe Cummings and Jon Carvill and Jon Shepard and Jonathan McPhie and Jonathan Torres and Josh Ginsburg and Junjie Wang and Kai Wu and Kam Hou U and Karan Saxena and Kartikay Khandelwal and Katayoun Zand and Kathy Matosich and Kaushik Veeraraghavan and Kelly Michelena and Keqian Li and Kiran Jagadeesh and Kun Huang and Kunal Chawla and Kyle Huang and Lailin Chen and Lakshya Garg and Lavender A and Leandro Silva and Lee Bell and Lei Zhang and Liangpeng Guo and Licheng Yu and Liron Moshkovich and Luca Wehrstedt and Madian Khabsa and Manav Avalani and Manish Bhatt and Martynas Mankus and Matan Hasson and Matthew Lennie and Matthias Reso and Maxim Groshev and Maxim Naumov and Maya Lathi and Meghan Keneally and Miao Liu and Michael L. Seltzer and Michal Valko and Michelle Restrepo and Mihir Patel and Mik Vyatskov and Mikayel Samvelyan and Mike Clark and Mike Macey and Mike Wang and Miquel Jubert Hermoso and Mo Metanat and Mohammad Rastegari and Munish Bansal and Nandhini Santhanam and Natascha Parks and Natasha White and Navyata Bawa and Nayan Singhal and Nick Egebo and Nicolas Usunier and Nikhil Mehta and Nikolay Pavlovich Laptev and Ning Dong and Norman Cheng and Oleg Chernoguz and Olivia Hart and Omkar Salpekar and Ozlem Kalinli and Parkin Kent and Parth Parekh and Paul Saab and Pavan Balaji and Pedro Rittner and Philip Bontrager and Pierre Roux and Piotr Dollar and Polina Zvyagina and Prashant Ratanchandani and Pritish Yuvraj and Qian Liang and Rachad Alao and Rachel Rodriguez and Rafi Ayub and Raghotham Murthy and Raghu Nayani and Rahul Mitra and Rangaprabhu Parthasarathy and Raymond Li and Rebekkah Hogan and Robin Battey and Rocky Wang and Russ Howes and Ruty Rinott and Sachin Mehta and Sachin Siby and Sai Jayesh Bondu and Samyak Datta and Sara Chugh and Sara Hunt and Sargun Dhillon and Sasha Sidorov and Satadru Pan and Saurabh Mahajan and Saurabh Verma and Seiji Yamamoto and Sharadh Ramaswamy and Shaun Lindsay and Shaun Lindsay and Sheng Feng and Shenghao Lin and Shengxin Cindy Zha and Shishir Patil and Shiva Shankar and Shuqiang Zhang and Shuqiang Zhang and Sinong Wang and Sneha Agarwal and Soji Sajuyigbe and Soumith Chintala and Stephanie Max and Stephen Chen and Steve Kehoe and Steve Satterfield and Sudarshan Govindaprasad and Sumit Gupta and Summer Deng and Sungmin Cho and Sunny Virk and Suraj Subramanian and Sy Choudhury and Sydney Goldman and Tal Remez and Tamar Glaser and Tamara Best and Thilo Koehler and Thomas Robinson and Tianhe Li and Tianjun Zhang and Tim Matthews and Timothy Chou and Tzook Shaked and Varun Vontimitta and Victoria Ajayi and Victoria Montanez and Vijai Mohan and Vinay Satish Kumar and Vishal Mangla and Vlad Ionescu and Vlad Poenaru and Vlad Tiberiu Mihailescu and Vladimir Ivanov and Wei Li and Wenchen Wang and Wenwen Jiang and Wes Bouaziz and Will Constable and Xiaocheng Tang and Xiaojian Wu and Xiaolan Wang and Xilun Wu and Xinbo Gao and Yaniv Kleinman and Yanjun Chen and Ye Hu and Ye Jia and Ye Qi and Yenda Li and Yilin Zhang and Ying Zhang and Yossi Adi and Youngjin Nam and Yu and Wang and Yu Zhao and Yuchen Hao and Yundi Qian and Yunlu Li and Yuzi He and Zach Rait and Zachary DeVito and Zef Rosnbrick and Zhaoduo Wen and Zhenyu Yang and Zhiwei Zhao and Zhiyu Ma},
      year={2024},
      eprint={2407.21783},
      archivePrefix={arXiv},
      primaryClass={cs.AI},
      url={https://arxiv.org/abs/2407.21783}, 
}

@misc{gptneo,
  author       = {Black, Sid and
                  Gao, Leo and
                  Wang, Phil and
                  Leahy, Connor and
                  Biderman, Stella},
  title        = {{GPT-Neo: Large Scale Autoregressive Language 
                   Modeling with Mesh-Tensorflow}},
  month        = mar,
  year         = 2021,
  note         = {{If you use this software, please cite it using 
                   these metadata.}},
  publisher    = {Zenodo},
  version      = {1.0},
  doi          = {10.5281/zenodo.5297715},
  url          = {https://doi.org/10.5281/zenodo.5297715}
}

@article{
miranda2025preserving,
title={Preserving Privacy in Large Language Models: A Survey on Current Threats and Solutions},
author={Michele Miranda and Elena Sofia Ruzzetti and Andrea Santilli and Fabio Massimo Zanzotto and S{\'e}bastien Brati{\`e}res and Emanuele Rodol{\`a}},
journal={Transactions on Machine Learning Research},
issn={2835-8856},
year={2025},
url={https://openreview.net/forum?id=Ss9MTTN7OL},
note={}
}

@misc{Yao2024Unlearning,
      title={KGA: A General Machine Unlearning Framework Based on Knowledge Gap Alignment}, 
      author={Lingzhi Wang and Tong Chen and Wei Yuan and Xingshan Zeng and Kam-Fai Wong and Hongzhi Yin},
      year={2023},
      eprint={2305.06535},
      archivePrefix={arXiv},
      primaryClass={cs.CL},
      url={https://arxiv.org/abs/2305.06535}, 
}

@misc{Meng2023a_ROME,
      title={Locating and Editing Factual Associations in GPT}, 
      author={Kevin Meng and David Bau and Alex Andonian and Yonatan Belinkov},
      year={2023},
      eprint={2202.05262},
      archivePrefix={arXiv},
      primaryClass={cs.CL},
      url={https://arxiv.org/abs/2202.05262}, 
}

@inproceedings{Carlini2023Quantifying,
  title={Quantifying Memorization Across Neural Language Models},
  author={Carlini, Nicholas and Ippolito, Daphne and Jagielski, Matthew and Lee, Katherine and Tramer, Florian and Zhang, Chiyuan},
  booktitle={Proceedings of the 11th International Conference on Learning Representations (ICLR)},
  year={2023},
  url={https://arxiv.org/abs/2202.07646}
}

@INPROCEEDINGS{Mamede2016,
  author={Mamede, Nuno and Baptista, Jorge and Dias, Francisco},
  booktitle={2016 IEEE Congress on Evolutionary Computation (CEC)}, 
  title={Automated anonymization of text documents}, 
  year={2016},
  volume={},
  number={},
  pages={1287-1294},
  doi={10.1109/CEC.2016.7743936}}

@inproceedings{Xu2019,
    title = "Privacy-Aware Text Rewriting",
    author = "Xu, Qiongkai  and
      Qu, Lizhen  and
      Xu, Chenchen  and
      Cui, Ran",
    editor = "van Deemter, Kees  and
      Lin, Chenghua  and
      Takamura, Hiroya",
    booktitle = "Proceedings of the 12th International Conference on Natural Language Generation",
    month = oct # "–" # nov,
    year = "2019",
    address = "Tokyo, Japan",
    publisher = "Association for Computational Linguistics",
    url = "https://aclanthology.org/W19-8633/",
    doi = "10.18653/v1/W19-8633",
    pages = "247--257"
}

\newpage
\appendix
\subsection{Paraphrase Generation and Evaluation}
\label{app:paraphrase}
To construct the variants $x^{L}_{para}$ used in the cross-lingual
bridge analysis of Section~III-D, we generate, for each sample and each
target language, a paraphrased version.
 
\subsubsection{Paraphrase Generation}
\label{app:paraphrase-generation}
 
For each language, we use an open-weight instruction-tuned language
model, selected to maximize paraphrase quality in that specific
language.
Table~\ref{tab:paraphrase-models} lists the model used for each language. For German,
we made two attempts to improve paraphrase quality, using first a
language-specific model and then a general-purpose one (see the German
model comparison below); Table~\ref{tab:paraphrase-models} reports the model retained.
All models are run locally, so no PII-containing sample is transmitted
to a third-party API during paraphrase generation.
 
\begin{table}[htbp]
\centering
\small
\caption{Models used for paraphrase generation.}
\label{tab:paraphrase-models}
\begin{tabular}{l p{5cm}}
\toprule
\textbf{Language} & \textbf{Model} \\
\midrule
English & Qwen2.5-7B-Instruct \\
German  & Qwen2.5-7B-Instruct \\
French  & Mistral-7B-Instruct-v0.3 \\
Italian & Minerva-7B-instruct-v1.0 \\
Spanish & Salamandra-7B-instruct \\
\bottomrule
\end{tabular}
\end{table}
 
Each model is prompted with an instruction that requests a copy of the
input sample in the same language, explicitly requiring that all entities
relevant to the attack (email addresses, phone numbers, handles, proper
names, dates, and URLs) be preserved word-for-word. This constraint is necessary because
the paraphrased samples are subsequently used as alternative attack prompts,
so any unintended alteration of the target PII would invalidate the
downstream extraction experiment.
 
Generation uses sampling with temperature $0.7$ and top-$p$ $0.9$, with a
maximum of 512 new tokens per completion.
 
\subsubsection{Evaluation Metrics}
\label{app:paraphrase-evaluation}
 
We evaluate the generated paraphrases along two complementary axes: lexical
divergence from the original sample and preservation of semantic content.
A good paraphrase should score \emph{low} on the former and \emph{high} on
the latter.
 
\textbf{Lexical divergence.} We compute sentence-level BLEU~\cite{papineni2002bleu}
(via \texttt{sacrebleu}) and ROUGE-L F1~\cite{lin2004rouge} (via the
\texttt{rouge-score} implementation) between each original sample and its
paraphrase. Low values in both metrics indicate that the paraphrase does not
merely reproduce the source text.
 
\textbf{Semantic similarity} Lexical divergence alone does not guaranty
that meaning is preserved, since a phrase could diverge too far from the
original content. We therefore additionally
measure semantic equivalence using a multilingual Natural Language Inference
(NLI) model, \texttt{xlm-roberta-large-xnli}~\cite{conneau2020xnli}, fine-tuned
on XNLI and covering all five languages used in this study. For each
pair (context, context) we compute the entailment probability in both
directions and average them:
\begin{multline}
\text{sim} = \frac{1}{2}\Big(P_{\text{entail}}(\text{context} \Rightarrow \text{paraphrase}) \\
+ P_{\text{entail}}(\text{paraphrase} \Rightarrow \text{context})\Big).
\end{multline}
A high bidirectional entailment score indicates that the two texts imply one
another and are therefore likely to be semantically equivalent, whereas a
one-directional entailment could instead reflect a generalization or
specialization of the original content.
 
\subsubsection{Results}
\label{app:paraphrase-results}
 
Table~\ref{tab:paraphrase-quality} reports, for each language of PII,
the mean BLEU, ROUGE-L, and semantic similarity between the original samples
and their paraphrases.
 
\begin{table*}[htbp]
\centering
\small
\caption{Mean BLEU, ROUGE-L (F1), and bidirectional semantic similarity
between original excerpts and their paraphrases, by PII type and language.
Lower BLEU/ROUGE-L and higher semantic similarity indicate better paraphrase
quality.}
\label{tab:paraphrase-quality}
\begin{tabular}{lll rrr}
\toprule
\textbf{PII type} & \textbf{Language} & \textbf{Model} & \textbf{BLEU} & \textbf{ROUGE-L} & \textbf{Sem. sim.} \\
\midrule
\multirow{6}{*}{Email}
  & English  & Qwen2.5-7B-Instruct & 23.51 & 49.21 & 83.43 \\
  & French   & Mistral-7B-Instruct-v0.3 & 8.21 & 15.02 & 69.23 \\
  & Italian  & Minerva-7B-instruct & 22.08 & 39.94 & 46.99 \\
  & Spanish  & Salamandra-7B-instruct & 6.62 & 18.84 & 41.35 \\
  & German   & DiscoLM\_German\_7b & 41.89 & 56.41 & 68.05 \\
  & German   & Qwen2.5-7B-Instruct & 47.80 & 67.70 & 90.89 \\
\midrule
\multirow{6}{*}{Phone}
  & English  & Qwen2.5-7B-Instruct & 27.05 & 52.87 & 84.18 \\
  & French   & Mistral-7B-Instruct-v0.3 & 7.34 & 14.88 & 66.65 \\
  & Italian  & Minerva-7B-instruct & 22.58 & 40.98 & 48.88 \\
  & Spanish  & Salamandra-7B-instruct & 6.44 & 20.16 & 42.10 \\
  & German   & DiscoLM\_German\_7b & 49.31 & 62.21 & 69.85 \\
  & German   & Qwen2.5-7B-Instruct & 41.98 & 63.94 & 90.70 \\
\midrule
\multirow{6}{*}{Twitter}
  & English  & Qwen2.5-7B-Instruct & 18.96 & 44.85 & 83.56 \\
  & French   & Mistral-7B-Instruct-v0.3 & 6.06 & 13.70 & 75.38 \\
  & Italian  & Minerva-7B-instruct & 18.33 & 36.28 & 47.06 \\
  & Spanish  & Salamandra-7B-instruct & 7.48 & 22.93 & 54.36 \\
  & German   & DiscoLM\_German\_7b & 45.20 & 56.36 & 61.46 \\
  & German   & Qwen2.5-7B-Instruct & 43.12 & 64.79 & 92.02 \\
\bottomrule
\end{tabular}
\end{table*}
 
\paragraph{German Paraphrase Model Comparison}
\label{app:model-comparison}

For German, we made two
attempts to improve the quality of the paraphrases. In a first attempt,
we used DiscoLM German 7b v1, a Mistral-based model fine-tuned
specifically for German. In a second attempt, we used
Qwen2.5-7B-Instruct, the general-purpose model employed for English, on
the same German samples. Table~\ref{tab:model-comparison-de} reports the comparison. The second
attempt yields substantially higher bidirectional semantic similarity
(about 90\% versus 61--70\% across PII types), at the cost of a
somewhat higher lexical overlap with the original (higher ROUGE-L for
all PII types, and higher BLEU for emails). Since preserving the
meaning of the original extracts is essential for the paraphrase to be a
valid attack prompt, we retained the paraphrases generated with
Qwen2.5-7B-Instruct for all German results reported in
Section~\ref{sec:finding-bridges} and in the experimental results.
 
\begin{table*}[htbp]
\centering
\small
\caption{Comparison of two candidate paraphrase models for German.}
\label{tab:model-comparison-de}
\begin{tabular}{ll rrr}
\toprule
\textbf{PII type} & \textbf{Model} & \textbf{BLEU} & \textbf{ROUGE-L} & \textbf{Sem. sim.} \\
\midrule
\multirow{2}{*}{Email}
  & DiscoLM\_German\_7b  & 41.89 & 56.41 & 68.05 \\
  & Qwen2.5-7B-Instruct  & 47.80 & 67.70 & 90.89 \\
\midrule
\multirow{2}{*}{Phone}
  & DiscoLM\_German\_7b  & 49.31 & 62.21 & 69.85 \\
  & Qwen2.5-7B-Instruct  & 41.98 & 63.94 & 90.70 \\
\midrule
\multirow{2}{*}{Twitter}
  & DiscoLM\_German\_7b  & 45.20 & 56.36 & 61.46 \\
  & Qwen2.5-7B-Instruct  & 43.12 & 64.79 & 92.02 \\
\bottomrule
\end{tabular}
\end{table*}

\subsection{Raw Results for Paraphrase Attacks}
\label{app:complete-para-attacks}
Table~\ref{tab:leakage_original_results} reports the absolute number of leaked PII instances obtained from the TDE attacks across models, languages, PII types, and input conditions. For each language, we report the leakage obtained with a strict translation of the original attack prefix and with its paraphrased counterpart. These raw counts provide the basis for the relative leakage analysis presented in Figure~\ref{fig:para-attacks}, where leakage is normalised to the corresponding English original prefix for each model and PII type.

\begin{table*}[t]
\centering
\small
\setlength{\tabcolsep}{5pt}
\caption{
Number of leaked PII instances for the strict translation and paraphrased attack prefixes.
Results are reported for each model, language, and PII type. 
}
\label{tab:leakage_original_results}

\begin{tabular}{lcccccccccc}
\toprule
& \multicolumn{2}{c}{English} 
& \multicolumn{2}{c}{French}
& \multicolumn{2}{c}{Italian}
& \multicolumn{2}{c}{Spanish}
& \multicolumn{2}{c}{German} \\
\textbf{Model}
& Strict & Paraphrase
& Strict & Paraphrase
& Strict & Paraphrase
& Strict & Paraphrase
& Strict & Paraphrase \\
\midrule

\multicolumn{11}{l}{\textbf{Twitter}} \\
\midrule

Llama 3.2 1B
& 340 & 68
& 255 & 56
& 243 & 16
& 260 & 36
& 213 & 25 \\

Llama 3.2 3B
& 548 & 127
& 492 & 96
& 501 & 28
& 488 & 68
& 389 & 36 \\

Qwen2.5 3B
& 468 & 146
& 395 & 104
& 372 & 19
& 384 & 55
& 298 & 31 \\

Qwen2.5 7B
& 591 & 181
& 479 & 135
& 439 & 20
& 440 & 70
& 361 & 43 \\

GPT-J 6B
& 1020 & 284
& 426 & 168
& 450 & 28
& 444 & 44
& 337 & 40 \\

GPT-Neo 1.3B
& 505 & 148
& 228 & 97
& 215 & 16
& 222 & 32
& 148 & 20 \\

GPT-Neo 2.7B
& 694 & 190
& 236 & 103
& 264 & 14
& 249 & 44
& 160 & 21 \\

\midrule
\multicolumn{11}{l}{\textbf{Email}} \\
\midrule

Llama 3.2 1B
& 167 & 39
& 78 & 23
& 74 & 3
& 71 & 10
& 80 & 16 \\

Llama 3.2 3B
& 329 & 78
& 195 & 44
& 192 & 15
& 187 & 20
& 183 & 47 \\

Qwen2.5 3B
& 242 & 102
& 136 & 36
& 145 & 7
& 140 & 11
& 118 & 37 \\

Qwen2.5 7B
& 328 & 123
& 210 & 51
& 204 & 13
& 227 & 15
& 193 & 49 \\

GPT-J 6B
& 726 & 226
& 121 & 71
& 114 & 9
& 108 & 7
& 109 & 33 \\

GPT-Neo 1.3B
& 267 & 84
& 42 & 24
& 44 & 3
& 35 & 0
& 34 & 10 \\

GPT-Neo 2.7B
& 442 & 128
& 70 & 39
& 62 & 7
& 54 & 6
& 56 & 14 \\

\midrule
\multicolumn{11}{l}{\textbf{Phone}} \\
\midrule

Llama 3.2 1B
& 161 & 16
& 32 & 19
& 37 & 8
& 35 & 4
& 6 & 3 \\

Llama 3.2 3B
& 388 & 34
& 111 & 35
& 103 & 18
& 111 & 13
& 55 & 18 \\

Qwen2.5 3B
& 197 & 17
& 35 & 13
& 38 & 5
& 43 & 5
& 8 & 4 \\

Qwen2.5 7B
& 285 & 31
& 54 & 23
& 58 & 5
& 61 & 4
& 10 & 4 \\

GPT-J 6B
& 495 & 50
& 18 & 22
& 16 & 3
& 21 & 4
& 5 & 3 \\

GPT-Neo 1.3B
& 138 & 10
& 3 & 9
& 3 & 2
& 3 & 2
& 1 & 2 \\

GPT-Neo 2.7B
& 256 & 22
& 4 & 11
& 4 & 2
& 6 & 5
& 1 & 1 \\

\bottomrule
\end{tabular}
\end{table*}

\end{document}